\documentclass[sigconf, noacm]{acmart}

\usepackage[english]{babel}

\usepackage{amsmath}
\usepackage{graphicx}
\usepackage{mathtools}
\usepackage{amsthm}
\usepackage{multirow}
\usepackage{breqn}
\usepackage{todonotes}
\usepackage{soul}
\usepackage{caption}
\usepackage{subcaption}
\usetikzlibrary{calc,matrix}

\usepackage[htt]{hyphenat}

\theoremstyle{acmplain}
\newtheorem{theorem}{Theorem}[section]
\newtheorem{proposition}[theorem]{Proposition}
\newtheorem{lemma}[theorem]{Lemma}

\theoremstyle{acmdefinition}
\newtheorem{definition}[theorem]{Definition}

\theoremstyle{remark}

\newcommand{\WG}[1]{\textcolor{black}{#1}}%
\newcommand{\NKT}[1]{\textcolor{black}{#1}} %
\newcommand{\LK}[1]{\textcolor{black}{#1}}%
\newcommand{\SW}[1]{\textcolor{black}{#1}}%
\newcommand{\LKb}[1]{\textcolor{black}{#1}}%

\DeclarePairedDelimiter\multiset{\lbrace\!\!\lbrace}{\rbrace\!\!\rbrace}
\DeclareMathOperator*{\clip}{clip}
\DeclareMathOperator{\Hash}{\text{\textsc{Hash}}}
\DeclareMathOperator{\Agg}{\text{\textsc{Aggregate}}}
\DeclareMathOperator{\Comb}{\text{\textsc{Combine}}}
\DeclareMathOperator{\ReLU}{\mathrm{ReLU}}
\DeclareMathOperator{\Readout}{\text{\textsc{ReadOut}}}

\DeclareMathOperator{\MLP}{\mathrm{MLP}}
\DeclareMathOperator{\Mean}{\mathrm{MEAN}}

\DeclareMathOperator{\Tree}{\text{\textsc{Tree}}}
\DeclareMathOperator*{\argmax}{arg\,max} %

\author{Lorenz Kummer}
\affiliation{%
  \institution{ Doctoral School Computer Science, Faculty of Computer Science, University of Vienna}
  \city{Vienna}
  \country{Austria}}
\email{lorenz.kummer@univie.ac.at}

\author{Samir Moustafa}
\affiliation{%
  \institution{Doctoral School Computer Science, Faculty of Computer Science, University of Vienna}
  \city{Vienna}
  \country{Austria}}
\email{samir.moustafa@univie.ac.at}

\author{Sebastian Schrittwieser}
\affiliation{%
  \institution{Christian Doppler Laboratory for Assurance and Transparency in Software Protection,\\ Faculty of Computer Science, University of Vienna}
  \city{Vienna}
  \country{Austria}}
\email{sebastian.schrittwieser@univie.ac.at}

\author{Wilfried Gansterer}
\affiliation{%
  \institution{Faculty of Computer Science, University of Vienna}
  \city{Vienna}
  \country{Austria}}
\email{wilfried.gansterer@univie.ac.at}

\author{Nils Kriege}
\affiliation{%
  \institution{Research Network Data Science, Faculty of Computer Science, University of Vienna} 
  \city{Vienna}
  \country{Austria}}
\email{nils.kriege@univie.ac.at}

\setcopyright{none} %
\acmConference[]{}{}{} %
\acmYear{} %
\acmISBN{} %
\acmDOI{} %
\renewcommand\footnotetextcopyrightpermission[1]{} %

\title[Attacking Graph Neural Networks with Bit Flips: Weisfeiler and \LK{Leman} Go Indifferent]{Attacking Graph Neural Networks with Bit Flips: \\
Weisfeiler and \LK{Leman} Go Indifferent}

\begin{document}
\begin{abstract}
\LKb{Prior} attacks on graph neural networks have focused on graph poisoning and evasion, neglecting the network's weights and biases. 
\NKT{
For convolutional neural networks, however, the risk arising from bit flip attacks is well recognized.
We show that the direct application of a traditional bit flip attack to graph neural networks is of limited effectivity.
Hence,} we
\WG{discuss} the Injectivity Bit Flip Attack, the first bit flip attack designed specifically for graph neural networks. 
\LKb{Our attack targets the learnable neighborhood aggregation functions in quantized message passing neural networks, degrading their ability to distinguish graph structures and
\WG{impairing} the expressivity of the Weisfeiler-Leman test.}
\LK{We find that} exploiting mathematical properties specific to certain graph neural networks %
\WG{significantly increases} 
their vulnerability to bit flip attacks. \WG{The} Injectivity Bit Flip Attack
can degrade the maximal expressive Graph Isomorphism Networks trained on %
graph property prediction datasets to random output by flipping only a small fraction of the network's bits, demonstrating
its higher destructive power compared to
\WG{traditional bit flip attacks} transferred from convolutional neural networks. %
\LK{Our attack is transparent, motivated by theoretical insights and 
confirmed by extensive 
empirical results.}
\end{abstract}
\maketitle

\section{Introduction}
\label{sec:intro}
Graph neural networks (GNNs) are a powerful machine learning technique for handling structured data represented as graphs with nodes and edges. These methods are highly versatile, extending the applicability of deep learning to new domains such as financial and social network analysis, medical data analysis, and chem- and bioinformatics~\citep{gao2022cancer,lu2021weighted, sun2021disease, xiong2021novo, cheung2020graph, wu2018moleculenet}.
With the increasing adoption of GNNs, there is a pressing need to investigate their potential security vulnerabilities. Traditional adversarial attacks on GNNs have focused on manipulating input graph data~\citep{wu2022graph} through poisoning attacks, which result in the learning of a faulty model~\citep{wu2022graph, ma2020towards}, or \WG{on} evasion attacks, which use adversarial examples to degrade inference. These attacks can be targeted~\citep{zunger2018adversarial} or untargeted~\citep{ma2020towards, zugner2019adversarial} and involve modifications
\WG{of} node features, edges, or the injection of new nodes~\citep{wu2022graph, sun2019node}. Targeted attacks degrade the model's performance on a subset of instances, while untargeted attacks degrade the model's overall performance~\citep{zhang2022unsupervised}. A classification of existing graph poisoning and evasion attacks and defense mechanisms as well as a repository with representative algorithms can be found in the comprehensive 
reviews by~\citet{jin2021adversarial} and~\citet{dai2022comprehensive}.

Previous research has shown that convolutional neural networks (CNNs) prominent in the computer vision domain are highly susceptible to Bit Flip Attacks (BFAs)~\citep{hong2019terminal, qian2023survey}. %
These works typically focus on CNNs to which quantization is applied, e.g.,~\citep{rakin2019bitflip, qian2023survey}, \LKb{a common technique to} improve efficiency. \LKb{Quantized CNNs are naturally more perturbation resistant compared to their unquantized (i.e., floating-point) counterparts~\citep{rakin2022tbfa, hong2019terminal}, and thus their degradation poses a more challenging problem. For example, quantization is mentioned by~\citet{hong2019terminal} as a viable protection mechanism against random BFAs.} \LKb{The perturbation resistance of quantized CNNs largely arises from their numerical representation. Specifically, using
\WG{integer representations} makes it difficult for a single bit flip to cause a significant increase \NKT{in absolute value}. 
For instance, in an 8-bit integer quantized format, the maximum increase a parameter can undergo due to a single bit flip is limited to 128. A flip in the exponent of an IEEE754 32-bit floating-point representation, however, can increase the parameter by \LKb{up to $~3.4 \cdot 10^{38}$}, %
causing extreme neuron activation overriding the rest of the activations~\citep{hong2019terminal, liu2017fault}.}
Efficient implementation is likewise crucial for practical applications of GNNs, making it necessary to investigate the interaction between robustness and efficiency. \LK{The practical relevance of quantizing GNNs has been recognized for applications such as inference on IoT and edge devices~\citep{yao2022edge}, reducing energy consumption for green deep learning~\citep{xu2021survey}, and in distributed GNN applications~\citep{shao2022distributed}. Recent efforts have been made to further advance quantization techniques for GNNs~\citep{zhu2023rm, bahri2021binary,feng2020sgquant} as well as technical realizations for their deployment~\citep{zhu2023rm, tailor2021degreequant} and further potential applications exist, e.g.,~\citep{dong2023graph, bert2023wireless, derrow2021eta}.} 

Prior research on the security of GNNs has \LKb{mostly overlooked} the potential \WG{threat} of BFAs~\citep{wu2022graph, jin2021adversarial, xu2020adversarial, ma2020towards} which directly manipulate a target model at inference time, and previous work on BFAs has not yet explored attacks on \LKb{quantized} GNNs~\citep{qian2023survey}. \LKb{Despite the known robustness of quantized CNNs to BFAs~\citep{rakin2022tbfa, hong2019terminal}, which potentially transfers to GNNs as it is mostly rooted in the numerical representation itself}, \LKb{the only
\WG{investigations of GNN} resilience to
\WG{BFAs} study bit flips in floating-point
\WG{representation}~\citep{wu2023securing, jiao2022assessing}.} Furthermore, while
general \LKb{BFAs}
\LKb{may be transferable} to GNNs, they do not consider
\WG{the} unique mathematical properties \WG{of GNNs}, although it has been observed that adapting BFAs to the specific properties of a target network can increase harm~\citep{venceslai2020neuroattack} and that BFAs can be far from optimal on non-convolutional models~\citep{hector2022evaluating}. 
We address this research gap by exploring the effects of malicious perturbations
\WG{of the trainable parameters of} \LKb{quantized} GNNs
and their impact on prediction quality. Specifically, we target the expressivity of GNNs, i.e., their ability to distinguish non-isomorphic graphs or node neighborhoods. The most expressive GNNs based on \NKT{standard} message passing, including the widely-used Graph Isomorphism Networks (GINs)~\citep{leskovec2019powerful}, have the same discriminative power as the 1-Weisfeiler-\LK{Leman} test (1-WL) for suitably parameterized neighborhood aggregation functions~\citep{wlsurvey, morris2019weisfeiler,leskovec2019powerful}. Our attack targets %
\LKb{the quantized parameters of these neighborhood aggregation functions} 
to degrade the network's ability to differentiate between non-isomorphic structures.

\subsection{Related work}
\label{ssec:problem}
The security issue of BFAs
\WG{has been} recognized for quantized CNNs, which are used in critical applications like medical image segmentation~\citep{zhang2021medq, askarihemmat2019u} and diagnoses~\citep{ribeiro2022ecg, garifulla2021case}. In contrast,
\WG{the robustness of} GNNs used in safety-critical domains, like medical diagnoses~\citep{gao2022cancer, lu2021weighted, li2020graph}, electronic health record modeling~\citep{sun2021disease, liu2020heterogeneous},
\WG{or} drug development~\citep{xiong2021novo, lin2020kgnn, cheung2020graph}, \WG{against BFAs}
\WG{has} not been sufficiently studied.
\citet{qian2023survey} and \citet{khare20222design} distinguish between targeted and untargeted BFAs similar to the distinction made between targeted and untargeted poisoning and evasion techniques.
The high volume of related work on BFAs for quantized CNNs~\citep{qian2023survey, khare20222design} based on the seminal work by~\citet{rakin2019bitflip}
and associated BFA defense mechanisms, e.g.,~\citep{khare20222design, li2021radar, liu2023neuropots}
published recently, underscores the need for research in the direction of both BFA and defense mechanisms for quantized GNNs.
\WG{As a representative for} these traditional BFA methods
\WG{we focus on} Progressive Bitflip Attack~\citep{rakin2019bitflip} as most other BFA variants are based on it. %
We subsequently refer to this specific BFA as \textbf{PBFA} and use the term BFA for the broader class of bit flip attacks only.
\LKb{Existing research on the resilience of GNNs to bit flips focuses exclusively on floating-point
\WG{representation}:~\citet{jiao2022assessing} study random bit faults and \citet{wu2023securing} consider a variant of %
\LKb{PBFA} modified to flip bits in the exponent of a weight's floating-point representation.} 

\subsection{Contribution}
\label{ssec:contri}
\LK{In a motivating case study, we explore \LKb{quantized} GNNs' vulnerability to PBFA, finding that PBFA does not notably surpass random bit flips in tasks demanding graph structural discrimination.}
\SW{To demonstrate that degrading \LKb{such} GNNs is nonetheless possible in the identified scenario,}
we introduce the Injectivity Bit Flip Attack (\textbf{IBFA}), a novel attack targeting the discriminative power of neighborhood aggregation functions used by GNNs. Specifically, we investigate the \NKT{GIN architecture with 1-WL expressivity}, where this function is injective for suitable parameters~\citep{leskovec2019powerful}, and which is integrated in popular frameworks~\citep{Fey/Lenssen/2019} and widely used in practice, e.g.,~\citep{bert2023wireless, wang2023dynamic, gao2022malware, yang2022identification}. %
\LK{IBFA, unlike existing BFAs for CNNs, has a unique bit-search algorithm optimization goal and input data selection strategy. It differs from graph poisoning and evasion attacks as it leaves input data unmodified. We establish a %
theoretical basis for IBFA,
\WG{and support it} by %
empirical evidence of its efficacy on real-world datasets. %
IBFA outperforms baselines in degrading prediction quality and requires fewer bit flips to %
reduce GIN’s output to randomness, making it indifferent to graph structures.}

\section{Preliminaries}
\label{sec:prelim}
\LK{IBFA targets quantized neighborhood aggregation-based GNNs, exploiting their expressivity linked to the 1-WL graph isomorphism test. We begin by summarizing these concepts %
and introduce our notation and terminology along the way.}
\label{ssec:notback}

\subsection{Graph theory}
A \emph{graph} $G$ is a pair $(V,E)$ of a finite set of \emph{nodes} $V$ and \emph{edges} $E \subseteq \left \{ \left \{u,v \right \} \subseteq V \right \}$. %
The set of nodes and edges of $G$ is denoted by $V(G)$ and $E(G)$, respectively. The \emph{neighborhood} of $v$ in $V(G)$ is $N(v) = \left \{ u \in V(G) \mid \{v, u\} \in E(G) \right \}$.
If there exists a bijection $\varphi \colon V(G) \rightarrow V(H)$ with $\{u,v\}$ in $E(G)$ if and only if $\{\varphi(u),  \varphi(v)\}$ in $E(H)$ for all $u$, $v$ in $V(G)$, we call the two graphs $G$ and $H$ \emph{isomorphic} and write $G \simeq H$. For two graphs with roots $r\in V(G)$ and $r'\in V(H)$, the bijection must additionally satisfy $\varphi(r)=r'$. The equivalence classes induced by $\simeq$ are referred to as \emph{isomorphism types}. 

A function $V(G) \rightarrow \Sigma$ is called a \emph{node coloring}. Then, a \emph{node colored} or \emph{labeled graph} $(G,l)$ is a graph $G$ endowed with a node coloring $l$. We call $l(v)$ a \emph{label} or \emph{color} of $v \in V(G)$. 
We denote a multiset by $\multiset{\dots}$.

\subsection{The Weisfeiler-\LK{Leman} algorithm}
\label{ssec:1wl}
Let $(G,l)$ denote a labelled graph. In every iteration $t > 0$, a node coloring $c_l^{(t)} \colon V(G) \rightarrow \Sigma$ is computed, which depends on the coloring $c_l^{(t-1)}$ of the previous iteration. At the beginning, the coloring is initialized as $c_l^{(0)} = l$. In subsequent iterations $t > 0$, the coloring is updated according to%
\begin{equation}
    \label{eq:wl_coloring}
    c_l^{(t)}(v) = \Hash\left ( c_l^{(t-1)}(v), \multiset{c_l^{(t-1)}(u)|u \in N(v)} \right ),
\end{equation}
  where $\Hash$ is an injective mapping of the above pair to a unique value in $\Sigma$, that has not been used in previous iterations. 
  The $\Hash$ function can, for example, be realized by assigning new consecutive integer values to pairs when they occur for the first time~\citep{shervashidze2011weisfeiler}.
Let $C_l^{(t)}(G)=\multiset{c_l^{(t)}(v) \mid v \in V(G)}$ be the multiset of colors a graph exhibits in iteration $t$.
The iterative coloring terminates if $|C_l^{(t-1)}(G)|=|C_l^{(t)}(G)|$, i.e., the number of colors does not change between two iterations. 
For testing whether two graphs $G$ and $H$ are isomorphic, the above algorithm is run in parallel on both $G$ and $H$.
If $C_l^{(t)}(G) \neq C_l^{(t)}(H)$ for any $t \geq 0$, then $G$ and $H$ are not isomorphic. 
The label assigned to a node $v$ in the $t$th iteration of the 1-WL test can be understood as a tree representation of the $t$-hop neighborhood of $v$, \LKb{in the sense that each $c_l^{(t)}(v)$ corresponds to an isomorphism type of
such trees of height $t$, see Definition~\ref{def:unfolding} and~\citep{dinverno2021aup,Jegelka2022GNNtheory, schulz2022weisfeiler} for details.}

\subsection{Graph neural networks}
\label{ssec:gnn}
Contemporary GNNs employ a neighborhood aggregation or message passing approach, in which the representation of a node is iteratively updated through the aggregation of representations of its neighboring nodes. 
Upon completion of $k$ iterations of aggregation, the representation of a node encapsulates the structural information within its $k$-hop neighborhood~\citep{leskovec2019powerful}. The $k$th layer of a GNN computes the node features $\mathbf{h}_v^{(k)}$ formally defined by
\begin{equation}
    \begin{aligned}
         \mathbf{a}_v^{(k)} &= \Agg^{(k)}\left ( \multiset{ \mathbf{h}_u^{(k-1)} \mid u \in N(v)} \right ),
         \\ 
         \mathbf{h}_v^{(k)} &= \Comb^{(k)}\left ( \mathbf{h}_v^{(k-1)}, \mathbf{a}_v^{(k)} \right ). \qquad
     \end{aligned}
    \label{eq:defgnn2}
\end{equation}
Initially, $\mathbf{h}_v^{(0)}$ are the \LK{graph's node features.}
\LKb{For graph level readout, individual node embeddings are aggregated into a single representation $\mathbf{h}_G$ of the entire graph}
\begin{equation}
    \mathbf{h}_G = \Readout \left ( \multiset{ \mathbf{h}_v^{(k)} \mid v \in G } \right ).
    \label{eq:defgnnreadout}
\end{equation} 
The choice of $\Agg^{(k)}$, $\Comb^{(k)}$ \LKb{and $\Readout$} in GNNs is critical, and several variants have been proposed~\citep{leskovec2019powerful}.

\subsection{Graph isomorphism network}
\label{sssec:gin}
\LK{GIN} has the same discriminative power as the 1-WL test in distinguishing non-isomorphic graphs~\citep{leskovec2019powerful}. A large body of work is devoted to GNNs exceeding this expressivity~\citep{wlsurvey}. However, neighborhood aggregation is widely used in practice and 1-WL is sufficient to distinguish most graphs in common benchmark datasets~\citep{Zopf22a, morris2021power}.
As established by~\citet{leskovec2019powerful}, a neighborhood aggregation GNN with a sufficient number of layers can reach the same discriminative power as the 1-WL test if both the $\Agg$ and $\Comb$ functions in each layer's update rule as well as its graph level $\Readout$ are injective. 
GIN achieves this with the update rule%
\begin{equation}
    \mathbf{h}_v^{(k)} = \MLP^{(k)}\left ( (1+\epsilon^{(k)} ) \cdot \mathbf{h}_v^{(k-1)} + \sum_{u \in N(v) } \mathbf{h}_u^{(k-1)} \right )
    \label{eq:defgin}
\end{equation}
integrating a multi layer perceptron (MLP) into  $\Comb^{(k)}$, realizing a universal function approximator on multisets~\citep{ZaheerKRPSS17, hornik1991approximation, hornik1989multilayer}.
If input features are one-hot encodings, an MLP is not needed before summation in the first layer, since summation is injective in this case. \LKb{Graph level readout in GIN is realized via first summing each layer's embeddings followed by concatenation (denoted by $\Vert$) of the summed vectors extracted from GINs' layers. The resulting graph level embedding $\mathbf{h}_G$ can then be used as input for, e.g., an $\MLP$ for subsequent graph level classifcation tasks.
\begin{equation}
    \mathbf{h}_G = \Big\Vert_{k=0}^n \left ( \sum_{ v \in V(G) } \mathbf{h}_v^{(k)} \right )
    \label{eq:defginreadout}
\end{equation} 
}
\subsection{Quantization}
\label{ssec:quant}
Quantization involves either a reduction in the precision of the weights, biases, and activations within a neural network or the use of a more efficient representation, resulting in a decrease in model size and memory utilization~\citep{adapt}. %
\label{sssec:quantmeth}
In accordance with the typical set-up chosen in the related work on BFA, see Section~\ref{ssec:problem}, we apply scale quantization to map \texttt{FLOAT32} tensors to the \texttt{INT8}
range.
\begin{equation}
    \begin{aligned}
        \mathcal{Q}(\mathbf{W}^{(l)}) &=  \mathbf{W}_{q}^{(l)} = \clip(\lfloor \mathbf{W}^{(l)} / s \rceil, \; a, b), 
        \\ 
        \mathcal{Q}^{-1}(\mathbf{W}_{q}^{(l)}) &= \widehat{\mathbf{W}}^{(l)} = \mathbf{W}_{q}^{(l)} \times s 
    \end{aligned}
    \label{equation:neural_network_qauntizer}
\end{equation}
specifies such a quantization function $\mathcal{Q}$ and its associated dequantization function $\mathcal{Q}^{-1}$. In Equation~\eqref{equation:neural_network_qauntizer}, $s$ is the scaling parameter, $\clip(x, a, b) = \min(\max(x,a),b)$ with $a$ and $b$ the \LKb{minimum} and \LKb{maximum} thresholds %
(also known as the quantization range), $\lfloor \dots \rceil$ denotes nearest integer rounding, $\mathbf{W}^{(l)}$ is the weight of a layer $l$ %
to be quantized, $\mathbf{W}_{q}^{(l)}$ its quantized counterpart and $\widehat{\dots}$ indicates a perturbation (i.e., rounding errors in the case of~\eqref{equation:neural_network_qauntizer}). %
Similar to other works on BFA that require quantized target networks, e.g.,~\citep{rakin2019bitflip}, we address the issue of non-differentiable rounding and clipping functions %
in $\mathcal{Q}$ by employing Straight Through Estimation (STE)~\citep{Bengio2013EstimatingOP}.

\subsection{Progressive bit flip attack}
\label{sssec:pbfa}
PBFA~\citep{rakin2019bitflip} uses a quantized trained CNN $\Phi$ and employs %
progressive bit search (\textbf{PBS}) to identify which bit flips will damage the CNN most. PBS begins with a single forward and backward pass, conducting error backpropagation without weight updates on a randomly selected batch $\mathbf{X}$ of training data with a target vector $\mathbf{t}$. It then selects the weights linked to the top-$k$ largest binary encoded gradients as potential candidates for flipping the associated bits. These candidate bits are iteratively tested (flipped) across all $L$ layers to find the bit that maximizes the difference between the loss $\mathcal{L}$ of the perturbed and the loss of the unperturbed CNN, %
whereby the same loss function is used that was minimized during training, e.g., (binary) cross entropy (CE) for (binary) classification. 
\begin{equation}
    \begin{aligned}
            \max_{\{ \widehat{\mathbf{W}}_{q}^{(l)} \}} \quad & \mathcal{L}\Big (\Phi \big( \mathbf{X}; \; \{\widehat{\mathbf{W}}_{q}^{(l)}\}_{l=1}^{L} \big), \; \mathbf{t} \Big) 
            - \mathcal{L}\Big (\Phi \big( \mathbf{X}; \; \{\mathbf{W}_{q}^{(l)}\}_{l=1}^{L} \big), \; \mathbf{t} \Big) \qquad
    \end{aligned}
    \label{eq:bfa_optimization}
\end{equation}
The source of the perturbation $\widehat{\dots}$ in Equation~\eqref{eq:bfa_optimization} are adversarial bit flips. %
If a single bit flip does not improve the optimization goal, PBS is executed again, considering combinations of two or more bit flips. This process continues until the desired level of network degradation is achieved or a predefined number of iterations is reached. Further details on PBFA/PBS can be found in the appendix.

\section{Injectivity bit flip attack} %
\label{sec:ibfa}
In principle, PBFA %
and potentially most other BFA variants based on it can be directly ported to GNNs. However,~\citet{hector2022evaluating} demonstrate that the high susceptibility of CNNs to BFA is closely tied to weight sharing in convolutional filters. \LK{Further,~\citet{hector2022evaluating} show that MLPs are inherently more robust to PBFA than CNNs due to their different gradient distributions.} The absence of convolutional filters in GNNs, \LK{as well as MLPs being an integral component of certain GNNs such as GIN}, motivates the development of a specialized attack for GNNs.

In our preliminary case study, %
(\LK{see} appendix), we examine PBFA's effectiveness on various GNN architectures. The case study's results indicate that quantized GINs trained on tasks requiring discrimination of graphs %
based on their structural properties and thus high structural expressivity as found in, e.g., drug development, display a remarkable resilience to PBFA, which in some instances hardly outperformed random bit flips \LK{even after more than $2\cdot10^3$ %
bit flips}. \LK{Considering contemporary literature~\citep{wang2023aegis, yao2020deephammer} postulates an upper limit of 24 to 50 precise bit flips to be achievable by an attacker within a span of several hours, the execution of such a substantial number of bit flips appears impractical.} 

Based on these observations, IBFA focuses on degrading GIN trained on tasks requiring high structural expressivity.
\label{ssec:theory}
The discriminative power of GIN is derived from its MLPs' ability to represent injective functions on sets 
(Section~\ref{sssec:gin}). Consequently, we design our attack \LK{assuming that} in 
tasks where learning such a discriminative function is crucial, attacking injectivity will lead to a higher degradation than performing PBFA. 

\subsection{Expressivity via injective set functions}
\label{par:inject}
A GNN based on message passing computing a function $F^{(k)}$ as output of the $k$th layer is maximal expressive if  
$F^{(k)}(u) = F^{(k)}(v) 
\Longleftrightarrow
c_l^{(k)}(u)=c_l^{(k)}(v)$. This is achieved when each layers' $\Comb$ and $\Agg$ functions are both injective, such that their combination is injective as well.
\begin{figure}%
    \begin{center}
    \centerline{\includegraphics[width=1.0\columnwidth, trim={0.025cm 0.075cm 0.38cm 0.17cm}, clip]{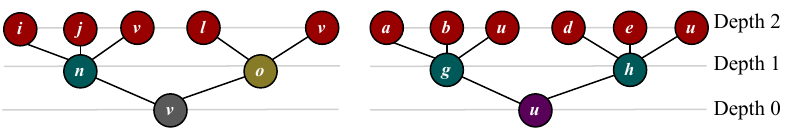}} 
    \caption{\label{fig:unfolding} Example of two non-isomorphic unfolding trees $T^{(2)}(u) \not \simeq T^{(2)}(v)$ of height 2 associated with the nodes $u$ and $v$. A function solving a WL-discriminitation task for $k=2$ must be able to discriminate $u$ and $v$ based on the structure of their unfolding trees.    }
    \end{center}
\end{figure} 
We develop the theory behind a bit flip attack exploiting this property.
We formally define the concept of an unfolding tree also known as \emph{computational tree} in the context of GNNs~\citep{Jegelka2022GNNtheory, dinverno2021aup}, see Figure~\ref{fig:unfolding} for an example.
\begin{definition}[Unfolding Tree~\citep{dinverno2021aup}]\label{def:unfolding}
The \emph{unfolding tree} $T^{(k)}(v)$ of height $k$ of a vertex $v$ is defined recursively as
\begin{equation*}
    T^{(k)}(v) = 
    \begin{cases}
    \Tree(l(v)) &\text{ if } k = 0,    \\ 
    \Tree(l(v), T^{(k-1)}(N(v))) & \text{ if } k > 0,
    \end{cases}
\end{equation*}
where $\Tree(l(v))$ is a tree containing a single node with label $l(v)$ and $\Tree(l(v), T^{(k-1)}(N(v)))$ is a tree with a root node labeled $l(v)$ having the roots of the trees $T^{(k-1)}(N(v)) = \{ T^{(k-1)}(w) \mid w \in N(v) \}$ as children.
\end{definition}
Unfolding trees are a convenient tool to study the expressivity of GNNs as they are closely related to the 1-WL colors.
\begin{lemma}[~\citep{dinverno2021aup}]\label{lem:wlc}
 Let $k \geq 0$ and $u$, $v$ nodes, then
 $c_l^{(k)}(u)=c_l^{(k)}(v) \Longleftrightarrow T^{(k)}(u) \simeq T^{(k)}(v).$
\end{lemma}
\citet{leskovec2019powerful} show that GIN can distinguish nodes with different WL colors. The result is obtained by arguing that the MLP in Equation~\eqref{eq:defgin} is a universal function approximator~\citep{hornik1989multilayer} allowing to learn arbitrary permutation invariant functions~\citep{ZaheerKRPSS17}. This includes, in particular, injective functions.
\LK{The complexity of GNNs regarding depth, width, and numerical precision required for this, however, remains poorly understood and is a focus of recent research~\citep{Aamand2022}.}

We investigate the functions \LK{of} a GIN layer and \LK{their contribution} to the expressivity of the final output, \LK{guiding the formulation of effective attacks}.
This requires determining whether our focus is on the expressivity of the general function on nodes or graphs \LK{in inductive learning or distinguishing} the elements of a predefined subset in \LK{transductive settings}.
First, we consider the general case, where a finite-depth GNN \LK{processes} all possible finite graphs and then discuss its implication for a concrete graph dataset. For the simplicity, we limit the discussion to unlabeled graphs.

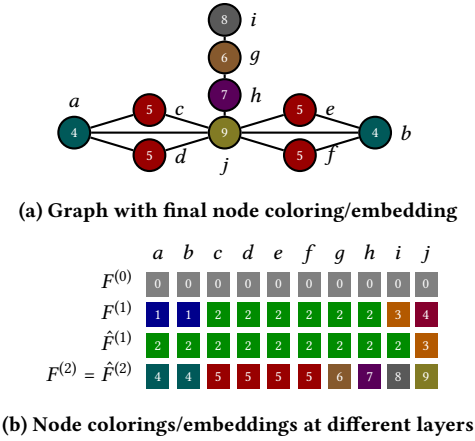
\begin{figure}[ht]\begin{center}
\colorlet{it0}{gray}

\colorlet{it11}{blue!55!black}
\colorlet{it12}{green!55!black}
\colorlet{it13}{orange!70!black}
\colorlet{it14}{purple!70!black}

\colorlet{it21}{red!60!black}
\colorlet{it22}{teal!70!black}
\colorlet{it23}{olive!80!black}
\colorlet{it24}{violet!70!black}
\colorlet{it25}{brown!70!black}
\colorlet{it26}{gray!70!black}

\begin{subfigure}[b]{1.0\columnwidth}\centering %
 \begin{tikzpicture}[scale=.5]
        \tikzset{edge/.style={draw,thick}}
        \tikzset{vertex/.style={edge,circle,inner sep=0pt,minimum width=12pt}}%
        \pgfdeclarelayer{background}
        \pgfdeclarelayer{foreground}
        \pgfsetlayers{background,main}

		\node [vertex,fill=it22,label={above:\small$a$}] (1) at (-4, 0) {\color{white}\tiny$4$};
		\node [vertex,fill=it21,label={right:\small$d$}] (2) at (-2, -.6) {\color{white}\tiny$5$};
		\node [vertex,fill=it21,label={right:\small$c$}] (3) at (-2, .6) {\color{white}\tiny$5$};
		\node [vertex,fill=it23,label={below:\small$j$}] (4) at (0,0) {\color{white}\tiny$9$};
		\node [vertex,fill=it21,label={right:\small$e$}] (5) at (2, .6) {\color{white}\tiny$5$};
		\node [vertex,fill=it21,label={right:\small$f$}] (6) at (2, -.6) {\color{white}\tiny$5$};
		\node [vertex,fill=it22,label={right:\small$b$}] (7) at (4, 0) {\color{white}\tiny$4$};
        \node [vertex,fill=it24,label={right:\small$h$}] (8) at (0, 1) {\color{white}\tiny$7$};
        \node [vertex,fill=it25,label={right:\small$g$}] (9) at (0, 2) {\color{white}\tiny$6$};
        \node [vertex,fill=it26,label={right:\small$i$}] (10) at (0, 3) {\color{white}\tiny$8$};

        \begin{pgfonlayer}{background}
		\draw[edge] (1) to (2);
		\draw[edge] (1) to (3);
		\draw[edge] (1) to (4);
		\draw[edge] (2) to (4);
		\draw[edge] (3) to (4);
		\draw[edge] (4) to (5);
		\draw[edge] (4) to (6);
		\draw[edge] (4) to (7);
		\draw[edge] (5) to (7);
		\draw[edge] (6) to (7);
        \draw[edge] (4) to (8);
        \draw[edge] (8) to (9);
        \draw[edge] (9) to (10);
        
        \end{pgfonlayer}
 \end{tikzpicture}  
 \caption{Graph with final node coloring/embedding}
 \label{fig:graph}
\end{subfigure}\hfill
\begin{subfigure}[b]{1.0\columnwidth}\centering%
\newcommand{\colcell}[3]{|[fill=#1]|{\color{#2}$#3$}}
\begin{tikzpicture}[ font=\tiny,
cell/.style={rectangle,draw=black},
space/.style={minimum height=0.9em,%
anchor=center,
minimum width=0.9em,%
matrix of nodes,
row sep=1.25pt,
column sep=1.25},
text depth=-0.1ex, 
nodes in empty cells,scale=.38, line width=1pt,
column 1/.style={anchor=base east}
]
\matrix [space] at (0,0){%
                      & \small$a$ & \small$b$ & \small$c$ & \small$d$ & \small$e$ & \small$f$ & \small$g$ & \small$h$ & \small$i$ & \small$j$ \\[-1.5mm]
\small$F^{(0)}$       &\colcell{it0}{white}{\large 0}&\colcell{it0}{white}{0}&\colcell{it0}{white}{0}&\colcell{it0}{white}{0}&\colcell{it0}{white}{0}&\colcell{it0}{white}{0}&\colcell{it0}{white}{0}&\colcell{it0}{white}{0}&\colcell{it0}{white}{0}&\colcell{it0}{white}{0} \\[-1mm]
\small$F^{(1)}$      &\colcell{it11}{white}{1}&\colcell{it11}{white}{1}&\colcell{it12}{white}{2}&\colcell{it12}{white}{2}&\colcell{it12}{white}{2}&\colcell{it12}{white}{2}&\colcell{it12}{white}{2}&\colcell{it12}{white}{2}&\colcell{it13}{white}{3}&\colcell{it14}{white}{4} \\[-1mm]
\small$\hat{F}^{(1)}$&\colcell{it12}{white}{2}&\colcell{it12}{white}{2}&\colcell{it12}{white}{2}&\colcell{it12}{white}{2}&\colcell{it12}{white}{2}&\colcell{it12}{white}{2}&\colcell{it12}{white}{2}&\colcell{it12}{white}{2}&\colcell{it12}{white}{2}&\colcell{it13}{white}{3} \\[-1mm]
\small$F^{(2)}=\hat{F}^{(2)}$       &\colcell{it22}{white}{4}&\colcell{it22}{white}{4}&\colcell{it21}{white}{5}&\colcell{it21}{white}{5}&\colcell{it21}{white}{5}&\colcell{it21}{white}{5}&\colcell{it25}{white}{6}&\colcell{it24}{white}{7}&\colcell{it26}{white}{8}&\colcell{it23}{white}{9} \\
};
\end{tikzpicture}
 \caption{Node colorings/embeddings at different layers}
 \label{fig:embeddings}
\end{subfigure}
 \caption{%
 \LK{Exemplary} results of the 2-layer GNNs $F^{(2)}$ and $\hat{F}^{(2)}$ using  $f^{(i)}$ and $\hat{f}^{(i)}$, respectively, for $i \in \{1,2\}$. Nodes having the same embedding are shown in the same color and are labeled with the same integer.
 Although $\hat{f}^{(1)}$ is non-injective and $\hat{F}^{(1)}$ is coarser than $F^{(1)}$, we have $F^{(2)}=\hat{F}^{(2)}$.
 The final output corresponds to the WL coloring.}
 \end{center}
\end{figure}

Let $f^{(i)}\colon \mathcal{M}(\mathbb{R}^{d_{i-1}}) \to \mathbb{R}^{d_i}$ be the learnable function of the $i$th layer of a GNN, where $\mathcal{M}(U)$ are all possible pairs $(A,\mathcal{A})$ with $A\in U$ and $\mathcal{A}$ a countable multisets of elements from $U$. We assume that $f^{(i)}$ is invariant regarding the order of elements in the multiset $\mathcal{A}$. 
Then the output of the network for node $v$ is obtained by the recursive function 
\begin{equation}\label{eq:recursive}
  F^{(k)}(v) = f^{(k)} \left(F^{(k-1)}(v), \multiset{F^{(k-1)}(w) \mid w \in N(v)}\right)
\end{equation}
with $F^{(0)}$ uniform initial node features.
Clearly, if all $f^{(i)}$ are injective, WL expressivity is reached as argued by~\citet{leskovec2019powerful}. 
The following proposition (proof in appendix) makes explicit that it suffices that all $f^{(i)}$ are injective with respect to the elements of their domain that represent (combinations of) unfolding trees of height $i-1$.
\begin{proposition}\label{thm:inject}
 Consider two arbitrary nodes $u$ and $v$ in an unlabeled graph.
 Let $\mathcal{J}_{0}$ be a uniform node feature and $\mathcal{J}_{i} = \{f^{(i)}(x) \mid x \in \mathcal{M}(\mathcal{J}_{i-1})\}$ the image under $f^{(i)}$ for $i>0$. Then
 \begin{equation}\label{eq:statementA}
   \forall i \leq k\colon\  \forall x,y \in \mathcal{M}(\mathcal{J}_{i-1})\colon\ f^{(i)}(x) = f^{(i)}(y) \Longrightarrow x = y 
 \end{equation}
   implies
 \begin{equation}\label{eq:statementB}
   c_l^{(k)}(u)=c_l^{(k)}(v) \Longleftrightarrow F^{(k)}(u) = F^{(k)}(v).
 \end{equation}
\end{proposition}
This result extends to graphs with discrete labels and continuous attributes.
\LK{The inputs, requiring distinct outputs from a GIN layer} to achieve WL expressivity, indicate weakpoints for potential attacks. These inputs are in 1-to-1 correspondence with the unfolding trees. \LK{As $i$ increases, the number of unfolding trees and the discriminative complexity for GIN's MLPs grows.} However, Proposition~\ref{thm:inject} provides only a sufficient condition for WL expressivity. \LK{Specifically, in a transductive setting on a concrete dataset, the variety of unfolding trees is limited by the dataset's node count. Additionally, even for a concrete dataset, the function $f^{(i)}$ at layer $i$ might be non-injective while $F^{(i+1)}$ remains maximally expressive, as illustrated in Figure~\ref{fig:embeddings}.}
This motivates the need for a targeted attack on injectivity to effectively degrade expressivity, which we develop below.
Further, these considerations lead to \LK{the following} exemplary classification tasks.

\subsection{Weisfeiler-Leman Discrimination Tasks}
\LKb{To develop an effective attack, we first need to provide a formulation of Prop.~\ref{thm:inject} that is more suitable to practical machine learning applications. Thus, we first introduce the following exemplary node level classification task, which we then extend to the graph level classification tasks which we will investigate experimentally.}
\begin{definition}[WL-discriminitation task]
\label{def:isodisc}
Let $G$ be a graph with labels $l$.
The \emph{WL-discriminitation task} for $k$ in $\mathbb{N}$ is to learn a function $F^{(k)}$ such that $F^{(k)}(u)=F^{(k)}(v)\Longleftrightarrow c_l^{(k)}(u) = c_l^{(k)}(v)$ for all $u,v \in V(G)$.
\end{definition}
The definition, which in its above form is concerned solely with node classification based on specific structural features, can easily be extended to classifying graphs based on their structure.

\LK{To define a graph level WL discrimination task, we first extend Equation~\eqref{eq:recursive} with a readout function $f_G \colon \mathcal{A} \rightarrow \mathbb{R}^{d_{G}}$, mapping a countable multiset $\mathcal{A}$ of elements from $U$ to a real valued vector representation of $G$:}
\begin{equation*}
    \LK{F^{(k)}_G(G) = f_G(\multiset{F^{(k)}(v) \mid v \in V(G)})}
\end{equation*}
\LK{Obviously, a GNN computing such a function is maximally expressive if $F^{(k)}_G(G_i)=F^{(k)}_G(G_j) \Longleftrightarrow C_l^{(k)}(G_i) = C_l^{(k)}(G_j)$ for all $G_i, G_j$. 
It further follows directly from $C_l^{(k)}(G)=\multiset{c_l^{(k)}(v) \mid v \in V(G)}$ that if the $k$th layer's message passing computation is maximally expressive, $c_l^{(k)}(u)=c_l^{(k)}(v) \Longleftrightarrow F^{(k)}(u) = F^{(k)}(v)$, and $f_G$'s injectivity is a sufficient condition for $F^{(k)}_G(G_i)=F^{(k)}_G(G_j) \Longleftrightarrow C_l^{(k)}(G_i) = C_l^{(k)}(G_j)$ to hold (Proposition~\ref{thm:inject2}, proof in appendix), which is consistent with results by~\citet{leskovec2019powerful}.} %
\begin{proposition}\label{thm:inject2}
\LK{Consider two arbitrary graphs $G_i, G_j$, 
 with $\mathcal{A}=\multiset{F^{(k)}(v) \mid v \in V(G_i)}$, $\mathcal{B}=\multiset{F^{(k)}(v) \mid v \in V(G_j)}$ and $c_l^{(k)}(u)=c_l^{(k)}(v) \Longleftrightarrow F^{(k)}(u) = F^{(k)}(v)$. Then
 \begin{equation}\label{eq:statementA2}
    f_G(\mathcal{A}) = f_G(\mathcal{B}) \Longrightarrow \mathcal{A} = \mathcal{B}
 \end{equation}
   implies
 \begin{equation}\label{eq:statementB2}
   F^{(k)}_G(G_i)=F^{(k)}_G(G_j) \Longleftrightarrow C_l^{(k)}(G_i) = C_l^{(k)}(G_j)
 \end{equation}}
\end{proposition}
\LKb{Note that analogous to Proposition~\ref{thm:inject}, the injectivity of $f_G$ with respect to the elements of its domain is only sufficient and not necessary for WL expressivity.}
\NKT{In particular, the sets $\mathcal{A}$ and $\mathcal{B}$ have a specific structure as they represent the unfolding trees of graphs. Hence, non-injective realization of $f_G$ might still suffice to distinguish graphs according to Equation~\eqref{eq:statementB2}.}
\begin{definition}[\LK{GLWL-discriminitation task}]
\label{def:isodisc2}
\LK{%
\LKb{Let $D$ be a set of node labeled graphs and let $G_i, G_j \in D$}. The \emph{Graph Level WL-discriminitation task} for $k$ in $\mathbb{N}$ is to learn a function $F^{(k)}_G$ such that $F^{(k)}_G(G_i)=F^{(k)}_G(G_j)$ if and only if $C_l^{(k)}(G_i) = C_l^{(k)}(G_j)$ for all $G_i, G_j \in D$ after $k$ iterations of the WL algorithm.} %
\end{definition}
\LK{According to Propositions~\ref{thm:inject} and~\ref{thm:inject2}, the injectivity of \( f^{(i)} \) and \( f_G \) is sufficient for achieving WL-level expressivity. Therefore, if a GNN has successfully learned a function \( F^{(k)}_G \) for a GLWL discrimination task, any BFA targeting the expressivity of such a GNN will likely compromise the injectivity of either \( f_G \) or \( f^{(i)} \).}

\LK{Real-world graph datasets, however, rarely satisfy the strict conditions outlined in Definition~\ref{def:isodisc} and~\ref{def:isodisc2} as \LKb{classification targets} used for supervised training typically do not perfectly align with the node's WL colors or the isomorphism \LK{types} of the dataset. Consequently, we formulate a relaxation of Definition~\ref{def:isodisc2} based on the Jaccard distance for multisets.} 
\begin{definition}[\LK{Multiset Jaccard distance~\citep{parapar2008winnow}}]
\label{def:jaccardm}
\LK{Let $A, B$ be multisets in universe $U$ and $m$ be the multiplicity function of multisets. \LKb{The \emph{multiset Jaccard distance} is then defined as $J_m(A,B) = 1 - \frac{\sum_{x \in U} \min(m_A(x), m_B(x))}{\sum_{x \in U} \max(m_A(x), m_B(x))}$, where $m_A(x)$ and $m_B(x)$ represent the multiplicity of element $x$ in multisets $A$ and $B$, respectively.}}
\end{definition}

\begin{definition}[\LK{$\varepsilon$-GLWL-discriminitation task}]
\label{def:isodisc4}
\LKb{%
\NKT{Let $D=\bigcup_{c\in C} D_c$  be a set of node labeled graphs with class labels $C$, where $D_c$ are the graphs of class $c \in C$.}
The \emph{$\varepsilon$-GLWL-discriminitation task} for $k$ in $\mathbb{N}$ is to learn a function $F^{(k)}_G$ such that for all $c \in C$ and all $G_i, G_j$ in $D_c=\{G_1,G_2,\dots,G_n\}$ it holds that $F^{(k)}_G(G_i)=F^{(k)}_G(G_j)$ if and only if $\frac{1}{\delta} \sum_{i=1}^{n-1} \sum_{j=i+1}^{n} J_m(C_l^{(k)}(G_i), C_l^{(k)}(G_j)) \leq \varepsilon$ for some $\varepsilon \in \mathbb{R}, 0 \leq \varepsilon < 1$ after $k$ iterations of the WL algorithm with $\delta=\binom{|D_c|}{2}$ denoting the number of unique pairs in $D_c$.}
\end{definition}
\LK{Definition~\ref{def:isodisc4} captures a broader trend regarding structural discrimination in the learning task, as it permits some graphs with the same classification target to exhibit \LKb{a degree (specified by $\varepsilon$) of} structural dissimilarity. Likewise, it permits graphs with different %
targets to show a degree of structural similarity.} \LKb{Nonetheless, any $\varepsilon < 1$ still requires $F_G^{(k)}$ to learn to distinguish certain substructures, for which injectivity of $f^{(k)}$ as well as $f_G$ with respect to an accordingly constrained domain remains a sufficient condition.}

\subsection{Targeting injectivity} %
It would not suffice to consider the injectivity of a single layer's $\Comb$ and $\Agg$ functions for an %
attack, as %
expressivity could be restored at deeper layers (\LK{Section~\ref{par:inject}, Figure~\ref{fig:embeddings}}). %
Thus, to target the injectivity \LKb{sufficent for} %
($\varepsilon$-GL)WL-discriminitation tasks as per Definition~\ref{def:isodisc}-~\ref{def:isodisc4} while considering the entire model, we reformulate the target of %
PBFA from a maximization Equation~\eqref{eq:bfa_optimization} to a minimization problem %
\begin{equation}
    \begin{aligned}
        \min_{\{ \widehat{\mathbf{W}}_{q}^{(l)} \}} \mathcal{L}\Big (\Phi \big( \mathbf{X}_a; \{\widehat{\mathbf{W}}_{q}^{(l)}\}_{l=1}^L \big),  \; \Phi\big ( \mathbf{X}_b;\{\widehat{\mathbf{W}}_{q}^{(l)}\}_{l=1}^L ) \Big).
    \end{aligned}
    \label{eq:optimibfa}
\end{equation}
That is, instead of increasing, e.g., the original classification loss of the model $\Phi$ via PBS, we use PBS to minimize the difference between the outputs of the network computed on two different inputs $\mathbf{X}_a$, and $\mathbf{X}_b$ w.r.t.\@ the function $\mathcal{L}$ that measures the difference between the network's outputs. \LK{This connects with our theoretical framework via ($\varepsilon$-GL)WL-discrimination tasks, in which \LKb{classification targets} (tend to) correlate with graph structural (dis)similarity.
Thus, inducing bit flips that exploit this tendency %
will potentially impair the injective mappings facilitated by the GNNs' $\Agg$ and $\Comb$ \LKb{functions}. %
Naturally, such an attack strategy necessaites careful selection of loss function and input data samples, which we discuss in detail below.}
This approach further allows us to perform IBFA on unlabeled data.

\subsubsection{Choosing the loss function}
\label{par:loss}
In a binary graph classification task, the network's outputs $\mathbf{y}_a = \Phi \big( \mathbf{X}_a; \{\widehat{\mathbf{W}}_{q}^{(l)}\}_{l=1}^L \big)$ as well as $\mathbf{y}_b = \Phi \big( \mathbf{X}_b; \{\widehat{\mathbf{W}}_{q}^{(l)}\}_{l=1}^L \big)$ both are $n \times 1$ vectors 
representing the probability mass functions (PMF) of $n$ Bernoulli distributed discrete random variables. For such distributed output vectors, any differentiable $p$-norm-based loss function would suffice to converge predictions in the sense of Equation~\eqref{eq:optimibfa} and we choose L1 for $\mathcal{L}$ for simplicity. In non-binary graph classification (i.e., multiclass-classification) or multitask binary classification, however, outputs $\mathbf{Y}_a$ and  $\mathbf{Y}_b$ are not $n \times 1$ vectors but instead $n \times m$ matrices where $n$ is the number of samples and $m$ the number of classes/tasks. That is, for each of the $n$ samples, each column in $\mathbf{Y}_a$ and  $\mathbf{Y}_b$ represents a PMF over $m$ classes. Thus, simply using a $p$-norm-based loss function as L1 for $\mathcal{L}$ in~\eqref{eq:optimibfa} would fail to capture differences in individual class probabilities contained in $\mathbf{Y}_a$ and $\mathbf{Y}_b$ due to the reduction operation required by L1 (e.g., mean or sum over $m$). We solve this by, instead of L1, employing the discrete pointwise Kullback-Leibler-Divergence~\citep{kullback1951information} (KL) as $\mathcal{L}$, i.e., the KL between the output's $n$ probability distributions of each pair of samples (data points) in $\mathbf{Y}_a$ and  $\mathbf{Y}_b$, which, in the context of~\eqref{eq:optimibfa}, allows IBFA to find %
bits %
converging the PMF of $\mathbf{Y}_a$ best on $\mathbf{Y}_b$. 
\begin{figure}
    \centering
    \includegraphics[width=1.0\columnwidth]{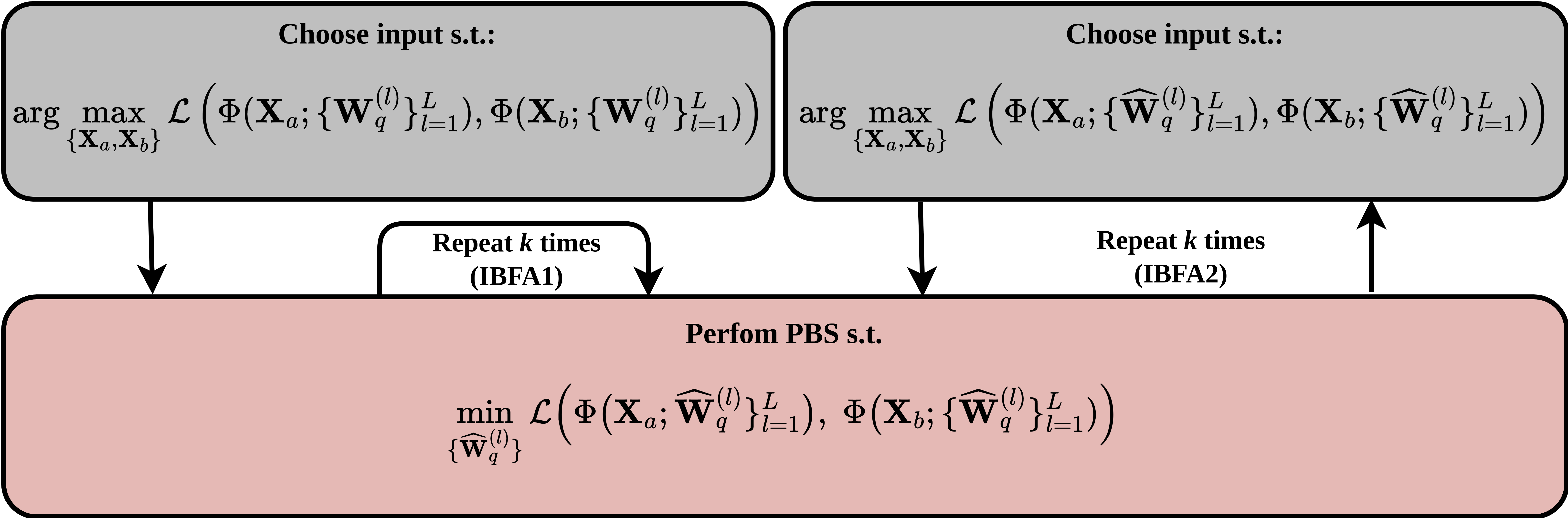}
    \caption{\label{fig:flowibfa}\LKb{IBFA1/2's integration of PBS %
    and input data selection strategies %
    for $k$ attack iterations. In the first attack iteration, input data selection of IBFA1 and IBFA2 are identical.}}
\end{figure}
\subsubsection{Choosing input samples}
\label{par:choosing}
The proper selection of $\mathbf{X}_a$ and $\mathbf{X}_b$ is crucial as selecting inputs that have identical outputs (e.g., two batches that contain different samples of the same classes in the same order) before the attack will not yield any degradation as Equation~\eqref{eq:optimibfa} would already be optimal.
\LK{Thus} we chose inputs $\mathbf{X}_a$ and $\mathbf{X}_b$ to be \LK{maximally} different from one another w.r.t.~to the unperturbed network's outputs by %
\begin{equation}\label{eq:find_sample}
    \argmax_{\{ \mathbf{X}_a, \mathbf{X}_b \}} \mathcal{L}\Big (\Phi \big( \mathbf{X}_a; \{\mathbf{W}_{q}^{(l)}\}_{l=1}^L \big),  \; \Phi\big ( \mathbf{X}_b;\{\mathbf{W}_{q}^{(l)}\}_{l=1}^L ) \Big).
\end{equation}
This search mechanism can be executed before the attack and the found $\mathbf{X}_a$, $\mathbf{X}_b$ reused for \LK{all attack iterations}, a variant of IBFA to which we refer as \textbf{IBFA1}. However, after \LK{each attack iteration}, the solution of~\eqref{eq:find_sample} might change, making it promising to recompute $\mathbf{X}_a$, $\mathbf{X}_b$ on the perturbed model before every subsequent attack iteration. We refer to the IBFA employing the latter data selection strategy as \textbf{IBFA2}. While IBFA2 may lead to slightly faster and more consistent degradation for a set amount of bit flips, %
its time complexity of $\Theta(k n^2)$ for $k$ attack runs and $n$ samples makes it less suitable for large datasets. \LKb{An overview of IBFA1 and IBFA2 is provided in Figure~\ref{fig:flowibfa}}.
\begin{figure*}%
    \begin{center}
    \centerline{\includegraphics[width=1.0\textwidth, trim={0.25cm 0.075cm 0.25cm 0.250cm}, clip]{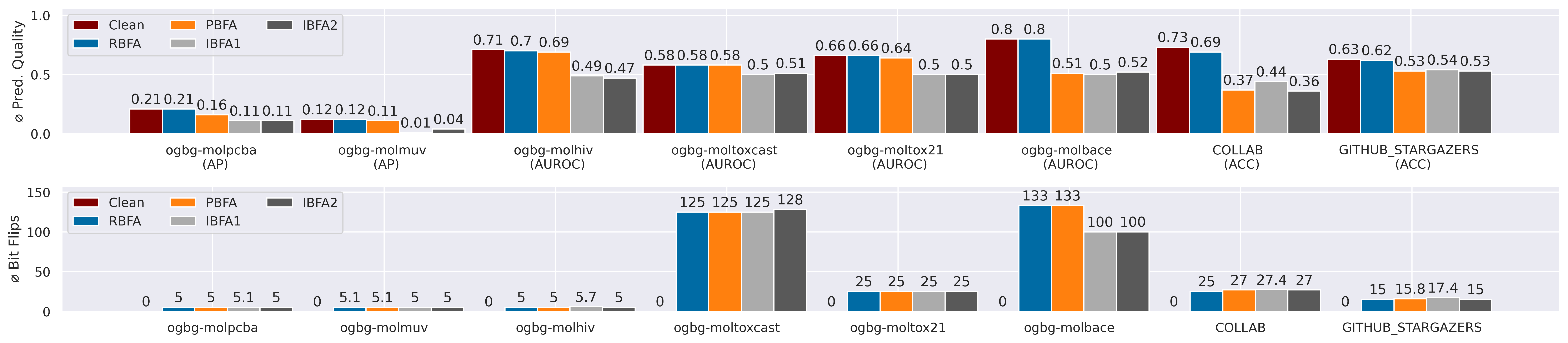}} 
    \caption{\label{fig:res} Pre- \LK{(clean)} and post-attack test quality metrics AP, AUROC or ACC for different BFA variants on a %
    5-layer GIN 
    trained on six \texttt{ogbg-mol} and two TUDataset %
    datasets, number of bit flips, averages of 10 runs.} %
    \end{center}
\end{figure*} 
\begin{figure*}%
    \begin{center}
    \centerline{\includegraphics[width=1.0\textwidth, trim={0.25cm 0.075cm 0.25cm 0.250cm}, clip]{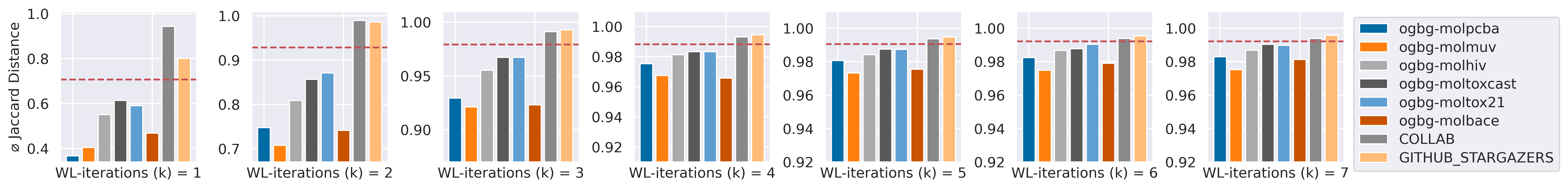}} 
    \caption{\label{fig:jaccard}\LK{\LKb{Jaccard distances} $J_m(C_l^{(k)}(G_i), C_l^{(k)}(G_j)), i \neq j$, averaged over all graphs $G_i, G_j$ and tasks (classes) in each randomly drawn sample of 3200 graphs per dataset, \LKb{indicating $\varepsilon$-GLWL-discriminitation tasks}. %
    The dotted lines distinguish molecular and social datasets for various numbers of WL-iterations ($k$) from 1 to 7 and \LKb{suggest} possible choices for $\varepsilon$.}} %
    \end{center}
\end{figure*} 

\subsection{Assumptions and threat model}
\label{ssec:limits}
\SW{Our work builds on several fundamental assumptions which are in line with previous work on BFA-based attacks on various types of neural networks.
Following %
previous} literature on BFAs for CNNs, we assume our target network is \texttt{INT8} quantized%
~\citep{liu2023neuropots, rakin2022tbfa, yao2020deephammer, rakin2019bitflip},
as such configured networks are naturally noise resistant~\citep{rakin2022tbfa}.

\SW{Furthermore,} we adopt the usual premise %
that the attacker has the capability to exactly flip the bits chosen by the bit-search algorithm through mechanisms such as RowHammer~\citep{mutlu2019rowhammer}, \LK{variants thereof~\citep{yao2020deephammer, lipp2020nethammer} or others~\citep{hou2020security, breier2018practical}}. %
The feasibility of \LK{BFAs} inducing exact bit flips via a RowHammer variant was shown by, e.g.,~\citep{wang2023aegis}. We thus do not \LK{discuss} the detailed technical specialities of realizing the flips of the identified vulnerable bits in hardware. %

\SW{Moreover, we assume a gray-box attack model in which the attacker's goal is to crush a trained and deployed quantized GNN via BFA. It represents a relaxation of some of the restrictions found in a black-box model by allowing the attacker to have certain prior knowledge about the training data and the general structure of the model. This is again a typical assumption also made in previous work~\citep{liu2023neuropots}. An attacker could easily obtain required information in many typical application scenarios, for instance, if the target model is a publicly available pre-trained model or the attacker has access to an identical instance of a device on which model inference is performed. Even in the absence of such a priori knowledge on the model, parameters and input data might be acquired through methods such as side-channel attacks~\citep{yan2020cache, batina2018csi}.} 

\section{Experiments}
\label{sec:experiments}
\LKb{Our experimental framework is designed to investigate the following key research questions, focusing on} real-world molecular property prediction datasets, a task common in drug development~\citep{xiong2021novo, rossi2020proximity}, as well as social network classifcation.
\LKb{
\begin{itemize}
    \item[\textbf{RQ1}] Is IBFA more destructive than other BFAs?%
    \item[\textbf{RQ2}] Does the destructiveness of IBFA depend on whether a task requires high structural expressivity (i.e., is an $\varepsilon$-GLWL-discrimination task)?
    \item[\textbf{RQ3}] Does IBFA's required number of bit flips fall within the accepted realistic budget of 24 to 50, as suggested in related work~\citep{wang2023aegis, yao2020deephammer}?
    \item[\textbf{RQ4}] How does IBFA perform in scenarios with limited data available for selection?
\end{itemize}
}
We \LKb{assess} IBFA \LKb{relative to} PBFA, which we consider the most relevant baseline,
as most other, more specialized (e.g., targeted) BFAs designed to degrade CNNs have been derived from PBFA.
We measured the degradation in the quality metrics proposed by Open Graph Benchmark (OGB)~\citep{hu2020open} or TUDataset~\citep{morris2020tudataset}, respectively for each of the datasets and followed the recommended variant of a 5-layers GIN with a virtual node. %
\LKb{For thoroughness,} we include a detailed ablation study concerning loss functions and layer preferences, %
which can be found in the appendix. \LKb{This study also} covers experiments on GNNs less expressive than GIN. To ensure reproducibility, we provide details on quantized models, measured metrics and attack configuration and a code repository\footnote{\url{https://github.com/lorenz0890/ibfakdd2024}}. %

\subsection{Quantized models}
We obtained \texttt{INT8} quantized models by training on each dataset's training split using STE, see Section~\ref{sssec:quantmeth}. We used the Adam optimizer with a learning rate of $10^{-3}$ and trained the models for 30 epochs. Although more complex models and quantization techniques might achieve higher prediction quality, our focus was not on improving prediction quality beyond the state-of-the-art, but on demonstrating GIN's vulnerability to IBFA. %
Some of the datasets we used present highly challenging learning tasks, %
and our results for quantized training of GIN are comparable to those %
by OGB~\citep{hu2020open} for \texttt{FLOAT32} training. %
\begin{figure*}%
    \centering %
    \includegraphics[width=2.0\columnwidth, trim={0.25cm 0.075cm 0.25cm 0.250cm}, clip]{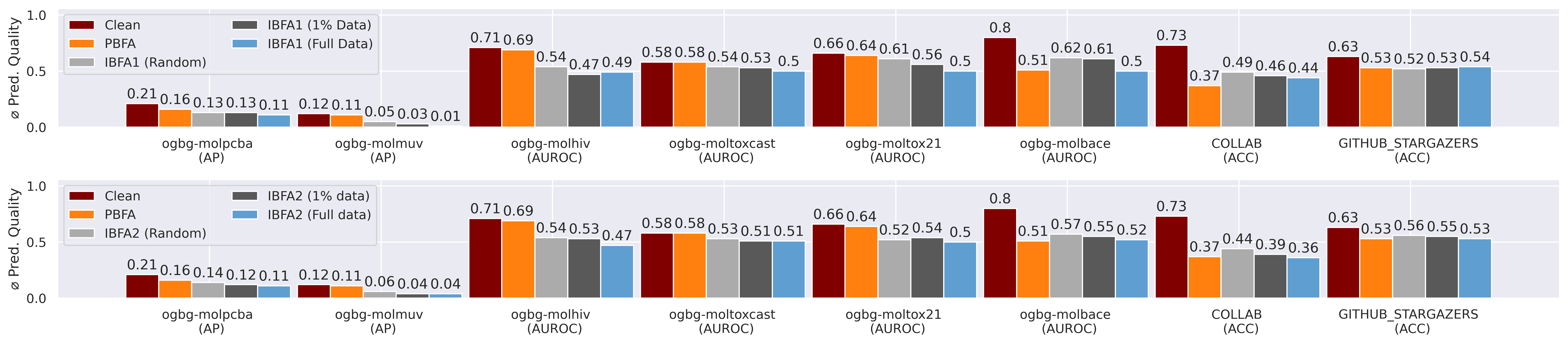} 
    \caption{\label{fig:res4} %
    Pre- (clean) and post-attack test quality metrics AP, AUROC or ACC for IBFA with different data selection strategies (random, IBFA selection from 1\% random subsets and full \LKb{training splits}) on a %
    5-layer GIN 
    trained on 6 \texttt{ogbg-mol} and 2 TUDataset %
    datasets, averages of 10 runs. IBFA1 in top row, IBFA2 in bottom row, number of bit flips as reported in Figure ~\ref{fig:res}.
    }
\end{figure*} 
\subsection{Datasets}
Six benchmark datasets are chosen (as in, e.g.,~\citep{gao2022patchgt, suresh2021adversarial}) from %
graph classification tasks from OGB based on MoleculeNet ~\citep{wu2018moleculenet} for evaluation as well as \texttt{COLLAB} and \texttt{GITHUB_STARGAZERS} from TUDataset~\citep{morris2020tudataset}. %
The goal in each OGB dataset is to predict properties based on molecular graph structures, such that the datasets are consistent with the underlying assumptions of IBFA described in Section~\ref{ssec:theory}.  All OGB datasets are split using a scaffold-based strategy, which %
seeks to separate structurally different molecules into different subsets~\citep{wu2018moleculenet, hu2020open}. \texttt{COLLAB} is derived from scientific
collaboration networks, whereby every graph represents the ego-network of a scientist, and the task is to predict their area of research. \texttt{GITHUB_STARGAZERS} contains graphs of social networks of GitHub users, and the task is to predict whether they starred popular machine learning or web development repositories. \texttt{COLLAB} and \texttt{GITHUB_STARGAZERS} are split randomly (80/10/10 for train/test/validation). More detailed dataset descriptions can be found in the appendix. 

Not all targets in the OGB datasets apply to each molecule (missing targets are indicated by NaNs) and we consider only existing targets in our experiments. %
Area under the receiver operating characteristic curve (AUROC), average precision (AP) or accuracy (ACC) are used to measure the models' performance %
as recommended by~\citet{hu2020open} and~\citet{morris2020tudataset}, respectively.

\subsection{Attack configuration}
\label{sssec:atackconfig}
The attacks in each of the experiments on a GIN trained on a dataset were executed with the number of attack runs (%
\LK{see} Section~\ref{sssec:pbfa}) initially set to $5$ and repeated with the number of attacks incremented until the first attack type reached (nearly) random output. The other attacks in this experiment were then set to that same number of attack runs to ensure fair comparison. Note that %
PBFA, IBFA1 and IBFA2 can flip more than a single bit during one attack run (\LK{see Section~\ref{sssec:pbfa})}, %
such that the final number of actual bit flips can vary across experiments. %
For the single task binary classification datasets, \texttt{ogbg-molhiv}, \texttt{ogbg-bace} and \texttt{GITHUB_STARGAZERS}, IBFA1/2 were used with $L1$ loss, for multitask binary classifcation \texttt{ogbg-tox21}, \texttt{ogbg-toxcast}, \texttt{ogbg-molmuv}, \texttt{ogbg-pcba} and multiclass classification (\texttt{COLLAB}), IBFA1/2 were used with KL loss. For PBFA, binary CE (BCE) loss was used throughout the binary classification datasets and CE loss was used for \texttt{COLLAB}. Input samples for all evaluated BFA variants were taken from the %
training splits.
\subsection{Results}
\label{ssec:results}
\begin{figure}[h]
    \centering %
    \includegraphics[width=1.0\columnwidth, trim={14.5cm 0.075cm 0.25cm 0.250cm}, clip]{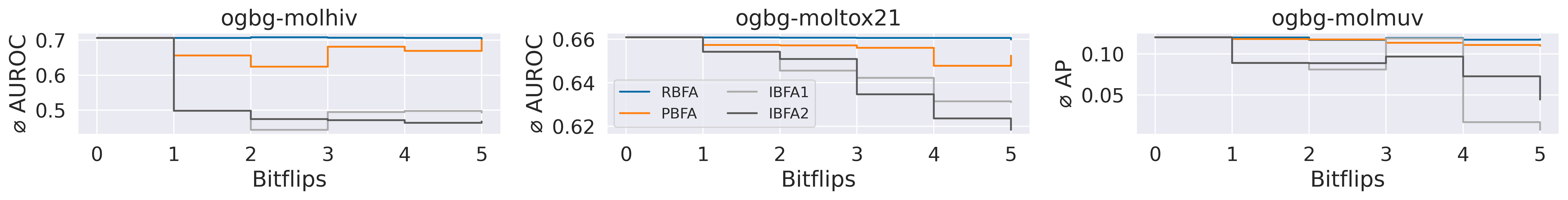} 
    \caption{\label{fig:res2} %
    Progression of degradation of a 5-layer GIN %
    trained on 2 small %
    datasets %
    with increasing total number of bit flips induced by different BFA variants, averages of 10 runs.}
\end{figure}

Both IBFA variants surpass random bit flips (\textbf{RBFA}) and PBFA in terms of test quality metric degradation for a given number of bit flips in most %
examined cases (Figure~\ref{fig:res}) \LKb{(\textbf{RQ1})}. \LKb{A fine granular visualization of the progression of quality metric degradation induced by IBFA1/2, PBFA and RBFA is given in Figure.~\ref{fig:res2}}. IBFA is capable of forcing the evaluated GINs to produce (almost) random output (AUROC $\leq 0.5$, AP $\leq 0.11$ (\texttt{ogbg-molpcba}), AP $\leq 0.06$ (\texttt{ogbg-molmuv}), ACC $\leq 0.33$ (\texttt{COLLAB}) or $\leq 0.5$ (\texttt{GITHUB_STARGAZERS})) by flipping less than 33 bits on average. \LK{This is \NKT{realizable given} the aforementioned upper limit of 24 to 50 precise bit flips an attacker can be expected to achieve%
~\citep{wang2023aegis, yao2020deephammer} \LKb{(\textbf{RQ3})}.} IBFA2 causes slightly \LK{more quality metric degradation} on \texttt{ogbg-molhiv}, \texttt{COLLAB} and \texttt{GITHUB_STARGAZERS} than IBFA1 but is surpassed by or on par with IBFA1 in all other cases. IBFA2 on \texttt{ogbg-molbace} and IBFA1 on \texttt{COLLAB} were slightly weaker than PBFA. \LK{However, on \texttt{ogbg-molbace} PBFA requires 33\% more bit flips to achieve results compareable to IBFA1.} On \texttt{GITHUB_STARGAZERS}, PBFA and IBFA both degraded GIN equally. %
GINs trained on \texttt{ogbg-molmuv}, \texttt{ogbg-molhiv}, and \texttt{ogbg-moltox21} were barely affected by PBFA for the examined number of bit flips and GIN trained on \texttt{ogbg-moltoxcast} appeared to be entirely impervious to PBFA. %
Our quantized GNNs %
resist RBFA, ruling out our %
observations %
are stochastic. %

\LK{These results are consistent with our definition of $\varepsilon$-GLWL-discriminitation tasks as Figure~\ref{fig:jaccard} indicates that, for suiteable choices of $\varepsilon$,
the examined \LKb{molecular graph property prediction datasets} from OGB %
\LKb{could} be distinguished from the \LKb{social network classification datasets} from TUDatset. That is, the OGB datasets demonstrate a stronger connection between class membership and structural graph similarity and thus, GINs trained on them are more vulnerable to IBFA. 
Together with our main results (Figure~\ref{fig:res}), \LKb{which show that PBFA either causes less degradation in the GINs trained on the OGB datasets than IBFA or requires more flips to do so}, we interpret these findings as empirical validation of our theoretical prediction that IBFA's destructiveness surpasses that of PBFA in tasks demanding high structural expressivity \LKb{(\textbf{RQ2})}.}
\LKb{The proposed data selection strategies for IBFA1/2 provide an improvement over random data selection (Figure~\ref{fig:res4}) and IBFA1/2 remain highly destructive (typically more destructive than PBFA), even when limited to a significantly constrained sample (1\% random subsets of the datasets' \LKb{training splits}) 
for selecting input data points (\textbf{RQ4}).}

\LK{%
\LKb{While our setup did not permit the weaker BFA in an experiment to continue flipping bits until the network was fully degraded}, our preliminary case study (appendix) revealed PBFA required 953 bit flips to fully degrade GIN on \texttt{ogbg-molhiv} and failed to degrade GIN on \texttt{ogbg-moltoxcast} even after 2662 bit flips. For comparison, IBFA1/2 could fully degrade GIN on both \texttt{ogbg-molhiv} and \texttt{ogbg-moltoxcast} using two orders of magnitude fewer bit flips (Figure~\ref{fig:res})}.

\section{Conclusion}
\label{sec:conc}
\LK{We introduce a %
\SW{novel concept} for bit flip attacks on \LKb{quantized} GNNs
\WG{which exploits} specific mathematical properties of GNNs
related to graph learning tasks requiring graph structural discrimination. Upon our theory, we
\WG{design the novel Injective Bit Flip Attack IBFA}
and illustrate its} ability to render GIN indifferent to graph structures.
\WG{IBFA compromises} its predictive quality %
significantly more than the most relevant BFA
ported from CNNs and \WG{than} random bit flips on eight molecular property prediction and social network classification datasets, covering binary and multiclass classification. %

\section*{Acknowledgements}
This work was supported by the Vienna Science and Technology Fund (WWTF) [10.47379/VRG19009]. The financial support by the Austrian Federal Ministry of Labour and Economy, the National Foundation for Research, Technology and Development and the Christian Doppler Research Association is gratefully acknowledged. We thank Tabea Reichmann for useful discussions. 

\bibliographystyle{ACM-Reference-Format}
\bibliography{kddbib}


\begin{thebibliography}{80}


\ifx \showCODEN    \undefined \def \showCODEN     #1{\unskip}     \fi
\ifx \showDOI      \undefined \def \showDOI       #1{#1}\fi
\ifx \showISBNx    \undefined \def \showISBNx     #1{\unskip}     \fi
\ifx \showISBNxiii \undefined \def \showISBNxiii  #1{\unskip}     \fi
\ifx \showISSN     \undefined \def \showISSN      #1{\unskip}     \fi
\ifx \showLCCN     \undefined \def \showLCCN      #1{\unskip}     \fi
\ifx \shownote     \undefined \def \shownote      #1{#1}          \fi
\ifx \showarticletitle \undefined \def \showarticletitle #1{#1}   \fi
\ifx \showURL      \undefined \def \showURL       {\relax}        \fi
\providecommand\bibfield[2]{#2}
\providecommand\bibinfo[2]{#2}
\providecommand\natexlab[1]{#1}
\providecommand\showeprint[2][]{arXiv:#2}

\bibitem[Aamand et~al\mbox{.}(2022)]%
        {Aamand2022}
\bibfield{author}{\bibinfo{person}{Anders Aamand}, \bibinfo{person}{Justin Chen}, \bibinfo{person}{Piotr Indyk}, \bibinfo{person}{Shyam Narayanan}, \bibinfo{person}{Ronitt Rubinfeld}, \bibinfo{person}{Nicholas Schiefer}, \bibinfo{person}{Sandeep Silwal}, {and} \bibinfo{person}{Tal Wagner}.} \bibinfo{year}{2022}\natexlab{}.
\newblock \showarticletitle{Exponentially Improving the Complexity of Simulating the Weisfeiler-Lehman Test with Graph Neural Networks}. In \bibinfo{booktitle}{\emph{Advances in Neural Information Processing Systems 35}}. \bibinfo{pages}{27333--27346}.
\newblock


\bibitem[Askari-Hemmat et~al\mbox{.}(2019)]%
        {askarihemmat2019u}
\bibfield{author}{\bibinfo{person}{Mohammad-Hossein Askari-Hemmat}, \bibinfo{person}{Sina Honari}, \bibinfo{person}{Lucas Rouhier}, \bibinfo{person}{Christian~S. Perone}, \bibinfo{person}{Julien Cohen-Adad}, \bibinfo{person}{Yvon Savaria}, {and} \bibinfo{person}{Jean-Pierre David}.} \bibinfo{year}{2019}\natexlab{}.
\newblock \showarticletitle{U-net fixed-point quantization for medical image segmentation}. In \bibinfo{booktitle}{\emph{Large-Scale Annotation of Biomedical Data and Expert Label Synthesis (LABELS) and Hardware Aware Learning for Medical Imaging and Computer Assisted Intervention (HAL-MICCAI), International Workshops}}. \bibinfo{pages}{115--124}.
\newblock


\bibitem[Bahri et~al\mbox{.}(2021)]%
        {bahri2021binary}
\bibfield{author}{\bibinfo{person}{Mehdi Bahri}, \bibinfo{person}{Ga{\'e}tan Bahl}, {and} \bibinfo{person}{Stefanos Zafeiriou}.} \bibinfo{year}{2021}\natexlab{}.
\newblock \showarticletitle{Binary graph neural networks}. In \bibinfo{booktitle}{\emph{Proceedings of the IEEE/CVF conference on computer vision and pattern recognition}}. \bibinfo{pages}{9492--9501}.
\newblock


\bibitem[Batina et~al\mbox{.}(2018)]%
        {batina2018csi}
\bibfield{author}{\bibinfo{person}{Lejla Batina}, \bibinfo{person}{Shivam Bhasin}, \bibinfo{person}{Dirmanto Jap}, {and} \bibinfo{person}{Stjepan Picek}.} \bibinfo{year}{2018}\natexlab{}.
\newblock \showarticletitle{CSI neural network: Using side-channels to recover your artificial neural network information}.
\newblock \bibinfo{journal}{\emph{CoRR}}  \bibinfo{volume}{abs/2204.07697} (\bibinfo{year}{2018}).
\newblock


\bibitem[Bengio et~al\mbox{.}(2013)]%
        {Bengio2013EstimatingOP}
\bibfield{author}{\bibinfo{person}{Yoshua Bengio}, \bibinfo{person}{Nicholas L{\'{e}}onard}, {and} \bibinfo{person}{Aaron~C. Courville}.} \bibinfo{year}{2013}\natexlab{}.
\newblock \showarticletitle{Estimating or Propagating Gradients Through Stochastic Neurons for Conditional Computation}.
\newblock \bibinfo{journal}{\emph{CoRR}}  \bibinfo{volume}{abs/1308.3432} (\bibinfo{year}{2013}).
\newblock


\bibitem[Bertalanič and Fortuna(2023)]%
        {bert2023wireless}
\bibfield{author}{\bibinfo{person}{Blaž Bertalanič} {and} \bibinfo{person}{Carolina Fortuna}.} \bibinfo{year}{2023}\natexlab{}.
\newblock \showarticletitle{Graph Isomorphism Networks for Wireless Link Layer Anomaly Classification}. In \bibinfo{booktitle}{\emph{2023 IEEE Wireless Communications and Networking Conference (WCNC)}}. \bibinfo{pages}{1--6}.
\newblock


\bibitem[Breier et~al\mbox{.}(2018)]%
        {breier2018practical}
\bibfield{author}{\bibinfo{person}{Jakub Breier}, \bibinfo{person}{Xiaolu Hou}, \bibinfo{person}{Dirmanto Jap}, \bibinfo{person}{Lei Ma}, \bibinfo{person}{Shivam Bhasin}, {and} \bibinfo{person}{Yang Liu}.} \bibinfo{year}{2018}\natexlab{}.
\newblock \showarticletitle{Practical Fault Attack on Deep Neural Networks}. In \bibinfo{booktitle}{\emph{Proceedings of the 2018 ACM SIGSAC Conference on Computer and Communications Security}}. \bibinfo{pages}{2204–2206}.
\newblock


\bibitem[Cheung and Moura(2020)]%
        {cheung2020graph}
\bibfield{author}{\bibinfo{person}{Mark Cheung} {and} \bibinfo{person}{Jos{\'e}~MF Moura}.} \bibinfo{year}{2020}\natexlab{}.
\newblock \showarticletitle{Graph neural networks for covid-19 drug discovery}. In \bibinfo{booktitle}{\emph{2020 IEEE International Conference on Big Data}}. \bibinfo{pages}{5646--5648}.
\newblock


\bibitem[Dai et~al\mbox{.}(2022)]%
        {dai2022comprehensive}
\bibfield{author}{\bibinfo{person}{Enyan Dai}, \bibinfo{person}{Tianxiang Zhao}, \bibinfo{person}{Huaisheng Zhu}, \bibinfo{person}{Junjie Xu}, \bibinfo{person}{Zhimeng Guo}, \bibinfo{person}{Hui Liu}, \bibinfo{person}{Jiliang Tang}, {and} \bibinfo{person}{Suhang Wang}.} \bibinfo{year}{2022}\natexlab{}.
\newblock \showarticletitle{A comprehensive survey on trustworthy graph neural networks: Privacy, robustness, fairness, and explainability}.
\newblock \bibinfo{journal}{\emph{CoRR}}  \bibinfo{volume}{abs/2204.08570} (\bibinfo{year}{2022}).
\newblock


\bibitem[Derrow-Pinion et~al\mbox{.}(2021)]%
        {derrow2021eta}
\bibfield{author}{\bibinfo{person}{Austin Derrow-Pinion}, \bibinfo{person}{Jennifer She}, \bibinfo{person}{David Wong}, \bibinfo{person}{Oliver Lange}, \bibinfo{person}{Todd Hester}, \bibinfo{person}{Luis Perez}, \bibinfo{person}{Marc Nunkesser}, \bibinfo{person}{Seongjae Lee}, \bibinfo{person}{Xueying Guo}, \bibinfo{person}{Brett Wiltshire}, {et~al\mbox{.}}} \bibinfo{year}{2021}\natexlab{}.
\newblock \showarticletitle{Eta prediction with graph neural networks in google maps}. In \bibinfo{booktitle}{\emph{Proceedings of the 30th ACM International Conference on Information \& Knowledge Management}}. \bibinfo{pages}{3767--3776}.
\newblock


\bibitem[Dong et~al\mbox{.}(2023)]%
        {dong2023graph}
\bibfield{author}{\bibinfo{person}{Guimin Dong}, \bibinfo{person}{Mingyue Tang}, \bibinfo{person}{Zhiyuan Wang}, \bibinfo{person}{Jiechao Gao}, \bibinfo{person}{Sikun Guo}, \bibinfo{person}{Lihua Cai}, \bibinfo{person}{Robert Gutierrez}, \bibinfo{person}{Bradford Campbel}, \bibinfo{person}{Laura~E Barnes}, {and} \bibinfo{person}{Mehdi Boukhechba}.} \bibinfo{year}{2023}\natexlab{}.
\newblock \showarticletitle{Graph neural networks in IoT: a survey}.
\newblock \bibinfo{journal}{\emph{ACM Transactions on Sensor Networks}} \bibinfo{volume}{19}, \bibinfo{number}{2} (\bibinfo{year}{2023}), \bibinfo{pages}{1--50}.
\newblock


\bibitem[D’Inverno et~al\mbox{.}(2021)]%
        {dinverno2021aup}
\bibfield{author}{\bibinfo{person}{Giuseppe~Alessio D’Inverno}, \bibinfo{person}{Monica Bianchini}, \bibinfo{person}{Maria~Lucia Sampoli}, {and} \bibinfo{person}{Franco Scarselli}.} \bibinfo{year}{2021}\natexlab{}.
\newblock \showarticletitle{A unifying point of view on expressive power of GNNs}.
\newblock \bibinfo{journal}{\emph{CoRR}}  \bibinfo{volume}{abs/2106.08992} (\bibinfo{year}{2021}).
\newblock


\bibitem[Feng et~al\mbox{.}(2020)]%
        {feng2020sgquant}
\bibfield{author}{\bibinfo{person}{Boyuan Feng}, \bibinfo{person}{Yuke Wang}, \bibinfo{person}{Xu Li}, \bibinfo{person}{Shu Yang}, \bibinfo{person}{Xueqiao Peng}, {and} \bibinfo{person}{Yufei Ding}.} \bibinfo{year}{2020}\natexlab{}.
\newblock \showarticletitle{SGQuant: Squeezing the Last Bit on Graph Neural Networks with Specialized Quantization}. In \bibinfo{booktitle}{\emph{2020 IEEE 32nd international conference on tools with artificial intelligence (ICTAI)}}. \bibinfo{pages}{1044--1052}.
\newblock


\bibitem[Fey and Lenssen(2019)]%
        {Fey/Lenssen/2019}
\bibfield{author}{\bibinfo{person}{Matthias Fey} {and} \bibinfo{person}{Jan~E. Lenssen}.} \bibinfo{year}{2019}\natexlab{}.
\newblock \showarticletitle{Fast Graph Representation Learning with {PyTorch Geometric}}. In \bibinfo{booktitle}{\emph{ICLR Workshop on Representation Learning on Graphs and Manifolds}}.
\newblock


\bibitem[Gao et~al\mbox{.}(2022a)]%
        {gao2022patchgt}
\bibfield{author}{\bibinfo{person}{Han Gao}, \bibinfo{person}{Xu Han}, \bibinfo{person}{Jiaoyang Huang}, \bibinfo{person}{Jian-Xun Wang}, {and} \bibinfo{person}{Liping Liu}.} \bibinfo{year}{2022}\natexlab{a}.
\newblock \showarticletitle{PatchGT: Transformer over Non-trainable Clusters for Learning Graph Representations}. In \bibinfo{booktitle}{\emph{Learning on Graphs Conference}}. \bibinfo{pages}{1--27}.
\newblock


\bibitem[Gao et~al\mbox{.}(2022c)]%
        {gao2022cancer}
\bibfield{author}{\bibinfo{person}{Jianliang Gao}, \bibinfo{person}{Tengfei Lyu}, \bibinfo{person}{Fan Xiong}, \bibinfo{person}{Jianxin Wang}, \bibinfo{person}{Weimao Ke}, {and} \bibinfo{person}{Zhao Li}.} \bibinfo{year}{2022}\natexlab{c}.
\newblock \showarticletitle{Predicting the Survival of Cancer Patients With Multimodal Graph Neural Network}.
\newblock \bibinfo{journal}{\emph{IEEE/ACM Transactions on Computational Biology and Bioinformatics}} \bibinfo{volume}{19}, \bibinfo{number}{2} (\bibinfo{year}{2022}), \bibinfo{pages}{699--709}.
\newblock


\bibitem[Gao et~al\mbox{.}(2022b)]%
        {gao2022malware}
\bibfield{author}{\bibinfo{person}{Yun Gao}, \bibinfo{person}{Hirokazu Hasegawa}, \bibinfo{person}{Yukiko Yamaguchi}, {and} \bibinfo{person}{Hajime Shimada}.} \bibinfo{year}{2022}\natexlab{b}.
\newblock \showarticletitle{Malware Detection by Control-Flow Graph Level Representation Learning With Graph Isomorphism Network}.
\newblock \bibinfo{journal}{\emph{IEEE Access}}  \bibinfo{volume}{10} (\bibinfo{year}{2022}), \bibinfo{pages}{111830--111841}.
\newblock


\bibitem[Garifulla et~al\mbox{.}(2021)]%
        {garifulla2021case}
\bibfield{author}{\bibinfo{person}{Mukhammed Garifulla}, \bibinfo{person}{Juncheol Shin}, \bibinfo{person}{Chanho Kim}, \bibinfo{person}{Won~Hwa Kim}, \bibinfo{person}{Hye~Jung Kim}, \bibinfo{person}{Jaeil Kim}, {and} \bibinfo{person}{Seokin Hong}.} \bibinfo{year}{2021}\natexlab{}.
\newblock \showarticletitle{A case study of quantizing convolutional neural networks for fast disease diagnosis on portable medical devices}.
\newblock \bibinfo{journal}{\emph{Sensors}} \bibinfo{volume}{22}, \bibinfo{number}{1} (\bibinfo{year}{2021}), \bibinfo{pages}{219}.
\newblock


\bibitem[Hector et~al\mbox{.}(2022)]%
        {hector2022evaluating}
\bibfield{author}{\bibinfo{person}{Kevin Hector}, \bibinfo{person}{Pierre-Alain Moëllic}, \bibinfo{person}{Mathieu Dumont}, {and} \bibinfo{person}{Jean-Max Dutertre}.} \bibinfo{year}{2022}\natexlab{}.
\newblock \showarticletitle{A Closer Look at Evaluating the Bit-Flip Attack Against Deep Neural Networks}. In \bibinfo{booktitle}{\emph{2022 IEEE 28th International Symposium on On-Line Testing and Robust System Design}}. \bibinfo{pages}{1--5}.
\newblock


\bibitem[Hong et~al\mbox{.}(2019)]%
        {hong2019terminal}
\bibfield{author}{\bibinfo{person}{Sanghyun Hong}, \bibinfo{person}{Pietro Frigo}, \bibinfo{person}{Yi{\u{g}}itcan Kaya}, \bibinfo{person}{Cristiano Giuffrida}, {and} \bibinfo{person}{Tudor Dumitraș}.} \bibinfo{year}{2019}\natexlab{}.
\newblock \showarticletitle{Terminal brain damage: Exposing the graceless degradation in deep neural networks under hardware fault attacks}. In \bibinfo{booktitle}{\emph{28th USENIX Security Symposium (USENIX Security 19)}}. \bibinfo{pages}{497--514}.
\newblock


\bibitem[Hornik(1991)]%
        {hornik1991approximation}
\bibfield{author}{\bibinfo{person}{Kurt Hornik}.} \bibinfo{year}{1991}\natexlab{}.
\newblock \showarticletitle{Approximation capabilities of multilayer feedforward networks}.
\newblock \bibinfo{journal}{\emph{Neural networks}} \bibinfo{volume}{4}, \bibinfo{number}{2} (\bibinfo{year}{1991}), \bibinfo{pages}{251--257}.
\newblock


\bibitem[Hornik et~al\mbox{.}(1989)]%
        {hornik1989multilayer}
\bibfield{author}{\bibinfo{person}{Kurt Hornik}, \bibinfo{person}{Maxwell Stinchcombe}, {and} \bibinfo{person}{Halbert White}.} \bibinfo{year}{1989}\natexlab{}.
\newblock \showarticletitle{Multilayer feedforward networks are universal approximators}.
\newblock \bibinfo{journal}{\emph{Neural networks}} \bibinfo{volume}{2}, \bibinfo{number}{5} (\bibinfo{year}{1989}), \bibinfo{pages}{359--366}.
\newblock


\bibitem[Hou et~al\mbox{.}(2020)]%
        {hou2020security}
\bibfield{author}{\bibinfo{person}{Xiaolu Hou}, \bibinfo{person}{Jakub Breier}, \bibinfo{person}{Dirmanto Jap}, \bibinfo{person}{Lei Ma}, \bibinfo{person}{Shivam Bhasin}, {and} \bibinfo{person}{Yang Liu}.} \bibinfo{year}{2020}\natexlab{}.
\newblock \showarticletitle{Security evaluation of deep neural network resistance against laser fault injection}. In \bibinfo{booktitle}{\emph{2020 IEEE International Symposium on the Physical and Failure Analysis of Integrated Circuits}}. \bibinfo{pages}{1--6}.
\newblock


\bibitem[Hu et~al\mbox{.}(2020)]%
        {hu2020open}
\bibfield{author}{\bibinfo{person}{Weihua Hu}, \bibinfo{person}{Matthias Fey}, \bibinfo{person}{Marinka Zitnik}, \bibinfo{person}{Yuxiao Dong}, \bibinfo{person}{Hongyu Ren}, \bibinfo{person}{Bowen Liu}, \bibinfo{person}{Michele Catasta}, {and} \bibinfo{person}{Jure Leskovec}.} \bibinfo{year}{2020}\natexlab{}.
\newblock \showarticletitle{Open graph benchmark: Datasets for machine learning on graphs}. In \bibinfo{booktitle}{\emph{Advances in neural information processing systems 33}}. \bibinfo{pages}{22118--22133}.
\newblock


\bibitem[Jegelka(2022)]%
        {Jegelka2022GNNtheory}
\bibfield{author}{\bibinfo{person}{Stefanie Jegelka}.} \bibinfo{year}{2022}\natexlab{}.
\newblock \showarticletitle{Theory of graph neural networks: Representation and learning}. In \bibinfo{booktitle}{\emph{Proceedings of the International Congress of Mathematicians}}, Vol.~\bibinfo{volume}{7}. \bibinfo{pages}{5450--5476}.
\newblock


\bibitem[Jiao et~al\mbox{.}(2022)]%
        {jiao2022assessing}
\bibfield{author}{\bibinfo{person}{Xun Jiao}, \bibinfo{person}{Ruixuan Wang}, \bibinfo{person}{Fred Lin}, \bibinfo{person}{Daniel Moore}, {and} \bibinfo{person}{Sriram Sankar}.} \bibinfo{year}{2022}\natexlab{}.
\newblock \showarticletitle{PyGFI: Analyzing and Enhancing Robustness of Graph Neural Networks Against Hardware Errors}.
\newblock \bibinfo{journal}{\emph{CoRR}}  \bibinfo{volume}{abs/2212.03475} (\bibinfo{year}{2022}).
\newblock


\bibitem[Jin et~al\mbox{.}(2021)]%
        {jin2021adversarial}
\bibfield{author}{\bibinfo{person}{Wei Jin}, \bibinfo{person}{Yaxing Li}, \bibinfo{person}{Han Xu}, \bibinfo{person}{Yiqi Wang}, \bibinfo{person}{Shuiwang Ji}, \bibinfo{person}{Charu Aggarwal}, {and} \bibinfo{person}{Jiliang Tang}.} \bibinfo{year}{2021}\natexlab{}.
\newblock \showarticletitle{Adversarial attacks and defenses on graphs}.
\newblock \bibinfo{journal}{\emph{ACM SIGKDD Explorations Newsletter}} \bibinfo{volume}{22}, \bibinfo{number}{2} (\bibinfo{year}{2021}), \bibinfo{pages}{19--34}.
\newblock


\bibitem[Khare et~al\mbox{.}(2022)]%
        {khare20222design}
\bibfield{author}{\bibinfo{person}{Yash Khare}, \bibinfo{person}{Kumud Lakara}, \bibinfo{person}{Maruthi~S. Inukonda}, \bibinfo{person}{Sparsh Mittal}, \bibinfo{person}{Mahesh Chandra}, {and} \bibinfo{person}{Arvind Kaushik}.} \bibinfo{year}{2022}\natexlab{}.
\newblock \showarticletitle{Design and Analysis of Novel Bit-flip Attacks and Defense Strategies for DNNs}. In \bibinfo{booktitle}{\emph{2022 IEEE Conference on Dependable and Secure Computing}}. \bibinfo{pages}{1--8}.
\newblock


\bibitem[Kullback and Leibler(1951)]%
        {kullback1951information}
\bibfield{author}{\bibinfo{person}{Solomon Kullback} {and} \bibinfo{person}{Richard Leibler}.} \bibinfo{year}{1951}\natexlab{}.
\newblock \showarticletitle{On information and sufficiency}.
\newblock \bibinfo{journal}{\emph{The annals of mathematical statistics}}  \bibinfo{volume}{22} (\bibinfo{year}{1951}), \bibinfo{pages}{79--86}.
\newblock


\bibitem[Kummer et~al\mbox{.}(2023)]%
        {adapt}
\bibfield{author}{\bibinfo{person}{Lorenz Kummer}, \bibinfo{person}{Kevin Sidak}, \bibinfo{person}{Tabea Reichmann}, {and} \bibinfo{person}{Wilfried Gansterer}.} \bibinfo{year}{2023}\natexlab{}.
\newblock \showarticletitle{Adaptive Precision Training (AdaPT): A dynamic quantized training approach for DNNs}. In \bibinfo{booktitle}{\emph{Proceedings of the 2023 SIAM International Conference on Data Mining}}. \bibinfo{pages}{559--567}.
\newblock


\bibitem[Li et~al\mbox{.}(2021)]%
        {li2021radar}
\bibfield{author}{\bibinfo{person}{Jingtao Li}, \bibinfo{person}{Adnan~Siraj Rakin}, \bibinfo{person}{Zhezhi He}, \bibinfo{person}{Deliang Fan}, {and} \bibinfo{person}{Chaitali Chakrabarti}.} \bibinfo{year}{2021}\natexlab{}.
\newblock \showarticletitle{RADAR: Run-time Adversarial Weight Attack Detection and Accuracy Recovery}. In \bibinfo{booktitle}{\emph{2021 Design, Automation and Test in Europe Conference and Exhibition}}. \bibinfo{pages}{790--795}.
\newblock


\bibitem[Li et~al\mbox{.}(2020)]%
        {li2020graph}
\bibfield{author}{\bibinfo{person}{Yang Li}, \bibinfo{person}{Buyue Qian}, \bibinfo{person}{Xianli Zhang}, {and} \bibinfo{person}{Hui Liu}.} \bibinfo{year}{2020}\natexlab{}.
\newblock \showarticletitle{Graph neural network-based diagnosis prediction}.
\newblock \bibinfo{journal}{\emph{Big Data}} \bibinfo{volume}{8}, \bibinfo{number}{5} (\bibinfo{year}{2020}), \bibinfo{pages}{379--390}.
\newblock


\bibitem[Lin et~al\mbox{.}(2020)]%
        {lin2020kgnn}
\bibfield{author}{\bibinfo{person}{Xuan Lin}, \bibinfo{person}{Zhe Quan}, \bibinfo{person}{Zhi-Jie Wang}, \bibinfo{person}{Tengfei Ma}, {and} \bibinfo{person}{Xiangxiang Zeng}.} \bibinfo{year}{2020}\natexlab{}.
\newblock \showarticletitle{KGNN: Knowledge Graph Neural Network for Drug-Drug Interaction Prediction}. In \bibinfo{booktitle}{\emph{Proceedings of the Twenty-Ninth International Joint Conference on Artificial Intelligence, {IJCAI-20}}}, Vol.~\bibinfo{volume}{380}. \bibinfo{pages}{2739--2745}.
\newblock


\bibitem[Lipp et~al\mbox{.}(2020)]%
        {lipp2020nethammer}
\bibfield{author}{\bibinfo{person}{Moritz Lipp}, \bibinfo{person}{Michael Schwarz}, \bibinfo{person}{Lukas Raab}, \bibinfo{person}{Lukas Lamster}, \bibinfo{person}{Misiker~Tadesse Aga}, \bibinfo{person}{Cl{\'e}mentine Maurice}, {and} \bibinfo{person}{Daniel Gruss}.} \bibinfo{year}{2020}\natexlab{}.
\newblock \showarticletitle{Nethammer: Inducing rowhammer faults through network requests}. In \bibinfo{booktitle}{\emph{2020 IEEE European Symposium on Security and Privacy Workshops}}. \bibinfo{pages}{710--719}.
\newblock


\bibitem[Liu et~al\mbox{.}(2023)]%
        {liu2023neuropots}
\bibfield{author}{\bibinfo{person}{Qi Liu}, \bibinfo{person}{Jieming Yin}, \bibinfo{person}{Wujie Wen}, \bibinfo{person}{Chengmo Yang}, {and} \bibinfo{person}{Shi Sha}.} \bibinfo{year}{2023}\natexlab{}.
\newblock \showarticletitle{{NeuroPots}: Realtime Proactive Defense against {Bit-Flip} Attacks in Neural Networks}. In \bibinfo{booktitle}{\emph{32nd USENIX Security Symposium (USENIX Security 23)}}. \bibinfo{pages}{6347--6364}.
\newblock


\bibitem[Liu et~al\mbox{.}(2017)]%
        {liu2017fault}
\bibfield{author}{\bibinfo{person}{Yannan Liu}, \bibinfo{person}{Lingxiao Wei}, \bibinfo{person}{Bo Luo}, {and} \bibinfo{person}{Qiang Xu}.} \bibinfo{year}{2017}\natexlab{}.
\newblock \showarticletitle{Fault injection attack on deep neural network}. In \bibinfo{booktitle}{\emph{2017 IEEE/ACM International Conference on Computer-Aided Design (ICCAD)}}. \bibinfo{pages}{131--138}.
\newblock


\bibitem[Liu et~al\mbox{.}(2020)]%
        {liu2020heterogeneous}
\bibfield{author}{\bibinfo{person}{Zheng Liu}, \bibinfo{person}{Xiaohan Li}, \bibinfo{person}{Hao Peng}, \bibinfo{person}{Lifang He}, {and} \bibinfo{person}{S~Yu Philip}.} \bibinfo{year}{2020}\natexlab{}.
\newblock \showarticletitle{Heterogeneous similarity graph neural network on electronic health records}. In \bibinfo{booktitle}{\emph{2020 IEEE International Conference on Big Data}}. \bibinfo{pages}{1196--1205}.
\newblock


\bibitem[Lu and Uddin(2021)]%
        {lu2021weighted}
\bibfield{author}{\bibinfo{person}{Haohui Lu} {and} \bibinfo{person}{Shahadat Uddin}.} \bibinfo{year}{2021}\natexlab{}.
\newblock \showarticletitle{A weighted patient network-based framework for predicting chronic diseases using graph neural networks}.
\newblock \bibinfo{journal}{\emph{Scientific reports}} \bibinfo{volume}{11}, \bibinfo{number}{1} (\bibinfo{year}{2021}), \bibinfo{pages}{22607}.
\newblock


\bibitem[Ma et~al\mbox{.}(2020)]%
        {ma2020towards}
\bibfield{author}{\bibinfo{person}{Jiaqi Ma}, \bibinfo{person}{Shuangrui Ding}, {and} \bibinfo{person}{Qiaozhu Mei}.} \bibinfo{year}{2020}\natexlab{}.
\newblock \showarticletitle{Towards More Practical Adversarial Attacks on Graph Neural Networks}. In \bibinfo{booktitle}{\emph{Advances in Neural Information Processing Systems 33}}. \bibinfo{pages}{4756--4766}.
\newblock


\bibitem[Morris et~al\mbox{.}(2021)]%
        {morris2021power}
\bibfield{author}{\bibinfo{person}{Christopher Morris}, \bibinfo{person}{Matthias Fey}, {and} \bibinfo{person}{Nils Kriege}.} \bibinfo{year}{2021}\natexlab{}.
\newblock \showarticletitle{The Power of the Weisfeiler-Leman Algorithm for Machine Learning with Graphs}. In \bibinfo{booktitle}{\emph{Proceedings of the Thirtieth International Joint Conference on Artificial Intelligence, {IJCAI-21}}}. \bibinfo{pages}{4543--4550}.
\newblock


\bibitem[Morris et~al\mbox{.}(2020)]%
        {morris2020tudataset}
\bibfield{author}{\bibinfo{person}{Christopher Morris}, \bibinfo{person}{Nils~M. Kriege}, \bibinfo{person}{Franka Bause}, \bibinfo{person}{Kristian Kersting}, \bibinfo{person}{Petra Mutzel}, {and} \bibinfo{person}{Marion Neumann}.} \bibinfo{year}{2020}\natexlab{}.
\newblock \showarticletitle{TUDataset: A collection of benchmark datasets for learning with graphs}. In \bibinfo{booktitle}{\emph{ICML 2020 Workshop on Graph Representation Learning and Beyond (GRL+ 2020)}}.
\newblock


\bibitem[Morris et~al\mbox{.}(2023)]%
        {wlsurvey}
\bibfield{author}{\bibinfo{person}{Christopher Morris}, \bibinfo{person}{Yaron Lipman}, \bibinfo{person}{Haggai Maron}, \bibinfo{person}{Bastian Rieck}, \bibinfo{person}{Nils~M. Kriege}, \bibinfo{person}{Martin Grohe}, \bibinfo{person}{Matthias Fey}, {and} \bibinfo{person}{Karsten Borgwardt}.} \bibinfo{year}{2023}\natexlab{}.
\newblock \showarticletitle{Weisfeiler and Leman go Machine Learning: The Story so far}.
\newblock \bibinfo{journal}{\emph{Journal of Machine Learning Research}} \bibinfo{volume}{24}, \bibinfo{number}{333} (\bibinfo{year}{2023}), \bibinfo{pages}{1--59}.
\newblock


\bibitem[Morris et~al\mbox{.}(2019)]%
        {morris2019weisfeiler}
\bibfield{author}{\bibinfo{person}{Christopher Morris}, \bibinfo{person}{Martin Ritzert}, \bibinfo{person}{Matthias Fey}, \bibinfo{person}{William~L. Hamilton}, \bibinfo{person}{Jan~Eric Lenssen}, \bibinfo{person}{Gaurav Rattan}, {and} \bibinfo{person}{Martin Grohe}.} \bibinfo{year}{2019}\natexlab{}.
\newblock \showarticletitle{Weisfeiler and leman go neural: Higher-order graph neural networks}. In \bibinfo{booktitle}{\emph{Proceedings of the AAAI conference on artificial intelligence}}, Vol.~\bibinfo{volume}{33}. \bibinfo{pages}{4602--4609}.
\newblock


\bibitem[Mutlu and Kim(2019)]%
        {mutlu2019rowhammer}
\bibfield{author}{\bibinfo{person}{Onur Mutlu} {and} \bibinfo{person}{Jeremie~S. Kim}.} \bibinfo{year}{2019}\natexlab{}.
\newblock \showarticletitle{Rowhammer: A retrospective}.
\newblock \bibinfo{journal}{\emph{IEEE Transactions on Computer-Aided Design of Integrated Circuits and Systems}} \bibinfo{volume}{39}, \bibinfo{number}{8} (\bibinfo{year}{2019}), \bibinfo{pages}{1555--1571}.
\newblock


\bibitem[Parapar and Barreiro(2008)]%
        {parapar2008winnow}
\bibfield{author}{\bibinfo{person}{Javier Parapar} {and} \bibinfo{person}{\'{A}lvaro Barreiro}.} \bibinfo{year}{2008}\natexlab{}.
\newblock \showarticletitle{Winnowing-Based Text Clustering}. In \bibinfo{booktitle}{\emph{Proceedings of the 17th ACM Conference on Information and Knowledge Management}}. \bibinfo{pages}{1353–1354}.
\newblock


\bibitem[Qian et~al\mbox{.}(2023)]%
        {qian2023survey}
\bibfield{author}{\bibinfo{person}{Cheng Qian}, \bibinfo{person}{Ming Zhang}, \bibinfo{person}{Yuanping Nie}, \bibinfo{person}{Shuaibing Lu}, {and} \bibinfo{person}{Huayang Cao}.} \bibinfo{year}{2023}\natexlab{}.
\newblock \showarticletitle{A Survey of Bit-Flip Attacks on Deep Neural Network and Corresponding Defense Methods}.
\newblock \bibinfo{journal}{\emph{Electronics}} \bibinfo{volume}{12}, \bibinfo{number}{4} (\bibinfo{year}{2023}), \bibinfo{pages}{853}.
\newblock


\bibitem[Rakin et~al\mbox{.}(2019)]%
        {rakin2019bitflip}
\bibfield{author}{\bibinfo{person}{Adnan~Siraj Rakin}, \bibinfo{person}{Zhezhi He}, {and} \bibinfo{person}{Deliang Fan}.} \bibinfo{year}{2019}\natexlab{}.
\newblock \showarticletitle{Bit-Flip Attack: Crushing Neural Network With Progressive Bit Search}. In \bibinfo{booktitle}{\emph{Proceedings of the IEEE/CVF International Conference on Computer Vision and Pattern Recognition}}. \bibinfo{pages}{1211--1220}.
\newblock


\bibitem[Rakin et~al\mbox{.}(2022)]%
        {rakin2022tbfa}
\bibfield{author}{\bibinfo{person}{Adnan~Siraj Rakin}, \bibinfo{person}{Zhezhi He}, \bibinfo{person}{Jingtao Li}, \bibinfo{person}{Fan Yao}, \bibinfo{person}{Chaitali Chakrabarti}, {and} \bibinfo{person}{Deliang Fan}.} \bibinfo{year}{2022}\natexlab{}.
\newblock \showarticletitle{T-BFA: Targeted Bit-Flip Adversarial Weight Attack}.
\newblock \bibinfo{journal}{\emph{IEEE Transactions on Pattern Analysis and Machine Intelligence}} \bibinfo{volume}{44}, \bibinfo{number}{11} (\bibinfo{year}{2022}), \bibinfo{pages}{7928--7939}.
\newblock


\bibitem[Ribeiro et~al\mbox{.}(2022)]%
        {ribeiro2022ecg}
\bibfield{author}{\bibinfo{person}{Henrique De~Melo Ribeiro}, \bibinfo{person}{Ahran Arnold}, \bibinfo{person}{James~P. Howard}, \bibinfo{person}{Matthew~J. Shun-Shin}, \bibinfo{person}{Ying Zhang}, \bibinfo{person}{Darrel~P. Francis}, \bibinfo{person}{Phang~B. Lim}, \bibinfo{person}{Zachary Whinnett}, {and} \bibinfo{person}{Massoud Zolgharni}.} \bibinfo{year}{2022}\natexlab{}.
\newblock \showarticletitle{ECG-based real-time arrhythmia monitoring using quantized deep neural networks: A feasibility study}.
\newblock \bibinfo{journal}{\emph{Computers in Biology and Medicine}}  \bibinfo{volume}{143} (\bibinfo{year}{2022}), \bibinfo{pages}{105249}.
\newblock


\bibitem[Rossi et~al\mbox{.}(2020)]%
        {rossi2020proximity}
\bibfield{author}{\bibinfo{person}{Ryan~A Rossi}, \bibinfo{person}{Di Jin}, \bibinfo{person}{Sungchul Kim}, \bibinfo{person}{Nesreen~K. Ahmed}, \bibinfo{person}{Danai Koutra}, {and} \bibinfo{person}{John~Boaz Lee}.} \bibinfo{year}{2020}\natexlab{}.
\newblock \showarticletitle{On proximity and structural role-based embeddings in networks: Misconceptions, techniques, and applications}.
\newblock \bibinfo{journal}{\emph{ACM Transactions on Knowledge Discovery from Data}} \bibinfo{volume}{14}, \bibinfo{number}{5} (\bibinfo{year}{2020}), \bibinfo{pages}{1--37}.
\newblock


\bibitem[Schulz et~al\mbox{.}(2022)]%
        {schulz2022weisfeiler}
\bibfield{author}{\bibinfo{person}{Till~Hendrik Schulz}, \bibinfo{person}{Tam\'{a}s Horv\'{a}th}, \bibinfo{person}{Pascal Welke}, {and} \bibinfo{person}{Stefan Wrobel}.} \bibinfo{year}{2022}\natexlab{}.
\newblock \showarticletitle{A Generalized Weisfeiler-Lehman Graph Kernel}.
\newblock \bibinfo{journal}{\emph{Machine Learning}} \bibinfo{volume}{111}, \bibinfo{number}{7} (\bibinfo{year}{2022}), \bibinfo{pages}{2601–2629}.
\newblock


\bibitem[Shao et~al\mbox{.}(2022)]%
        {shao2022distributed}
\bibfield{author}{\bibinfo{person}{Yingxia Shao}, \bibinfo{person}{Hongzheng Li}, \bibinfo{person}{Xizhi Gu}, \bibinfo{person}{Hongbo Yin}, \bibinfo{person}{Yawen Li}, \bibinfo{person}{Xupeng Miao}, \bibinfo{person}{Wentao Zhang}, \bibinfo{person}{Bin Cui}, {and} \bibinfo{person}{Lei Chen}.} \bibinfo{year}{2022}\natexlab{}.
\newblock \showarticletitle{Distributed graph neural network training: A survey}.
\newblock \bibinfo{journal}{\emph{CoRR}}  \bibinfo{volume}{abs/2211.00216} (\bibinfo{year}{2022}).
\newblock


\bibitem[Shervashidze et~al\mbox{.}(2011)]%
        {shervashidze2011weisfeiler}
\bibfield{author}{\bibinfo{person}{Nino Shervashidze}, \bibinfo{person}{Pascal Schweitzer}, \bibinfo{person}{Erik~Jan Van~Leeuwen}, \bibinfo{person}{Kurt Mehlhorn}, {and} \bibinfo{person}{Karsten~M Borgwardt}.} \bibinfo{year}{2011}\natexlab{}.
\newblock \showarticletitle{Weisfeiler-lehman graph kernels}.
\newblock \bibinfo{journal}{\emph{Journal of Machine Learning Research}} \bibinfo{volume}{12}, \bibinfo{number}{9} (\bibinfo{year}{2011}).
\newblock


\bibitem[Sun et~al\mbox{.}(2020)]%
        {sun2019node}
\bibfield{author}{\bibinfo{person}{Yiwei Sun}, \bibinfo{person}{Suhang Wang}, \bibinfo{person}{Xianfeng Tang}, \bibinfo{person}{Tsung-Yu Hsieh}, {and} \bibinfo{person}{Vasant Honavar}.} \bibinfo{year}{2020}\natexlab{}.
\newblock \showarticletitle{Adversarial Attacks on Graph Neural Networks via Node Injections: A Hierarchical Reinforcement Learning Approach}. In \bibinfo{booktitle}{\emph{Proceedings of The Web Conference 2020}}. \bibinfo{pages}{673–683}.
\newblock


\bibitem[Sun et~al\mbox{.}(2021)]%
        {sun2021disease}
\bibfield{author}{\bibinfo{person}{Zhenchao Sun}, \bibinfo{person}{Hongzhi Yin}, \bibinfo{person}{Hongxu Chen}, \bibinfo{person}{Tong Chen}, \bibinfo{person}{Lizhen Cui}, {and} \bibinfo{person}{Fan Yang}.} \bibinfo{year}{2021}\natexlab{}.
\newblock \showarticletitle{Disease Prediction via Graph Neural Networks}.
\newblock \bibinfo{journal}{\emph{IEEE Journal of Biomedical and Health Informatics}} \bibinfo{volume}{25}, \bibinfo{number}{3} (\bibinfo{year}{2021}), \bibinfo{pages}{818--826}.
\newblock


\bibitem[Suresh et~al\mbox{.}(2021)]%
        {suresh2021adversarial}
\bibfield{author}{\bibinfo{person}{Susheel Suresh}, \bibinfo{person}{Pan Li}, \bibinfo{person}{Cong Hao}, {and} \bibinfo{person}{Jennifer Neville}.} \bibinfo{year}{2021}\natexlab{}.
\newblock \showarticletitle{Adversarial graph augmentation to improve graph contrastive learning}. In \bibinfo{booktitle}{\emph{Advances in Neural Information Processing Systems 34}}. \bibinfo{pages}{15920--15933}.
\newblock


\bibitem[Tailor et~al\mbox{.}(2021)]%
        {tailor2021degreequant}
\bibfield{author}{\bibinfo{person}{Shyam~A. Tailor}, \bibinfo{person}{Javier Fernandez-Marques}, {and} \bibinfo{person}{Nicholas~D. Lane}.} \bibinfo{year}{2021}\natexlab{}.
\newblock \showarticletitle{Degree-Quant: Quantization-Aware Training for Graph Neural Networks}. In \bibinfo{booktitle}{\emph{9th International Conference on Learning Representations}}.
\newblock


\bibitem[Venceslai et~al\mbox{.}(2020)]%
        {venceslai2020neuroattack}
\bibfield{author}{\bibinfo{person}{Valerio Venceslai}, \bibinfo{person}{Alberto Marchisio}, \bibinfo{person}{Ihsen Alouani}, \bibinfo{person}{Maurizio Martina}, {and} \bibinfo{person}{Muhammad Shafique}.} \bibinfo{year}{2020}\natexlab{}.
\newblock \showarticletitle{Neuroattack: Undermining spiking neural networks security through externally triggered bit-flips}. In \bibinfo{booktitle}{\emph{2020 International Joint Conference on Neural Networks}}. \bibinfo{pages}{1--8}.
\newblock


\bibitem[Wang et~al\mbox{.}(2023b)]%
        {wang2023aegis}
\bibfield{author}{\bibinfo{person}{Jialai Wang}, \bibinfo{person}{Ziyuan Zhang}, \bibinfo{person}{Meiqi Wang}, \bibinfo{person}{Han Qiu}, \bibinfo{person}{Tianwei Zhang}, \bibinfo{person}{Qi Li}, \bibinfo{person}{Zongpeng Li}, \bibinfo{person}{Tao Wei}, {and} \bibinfo{person}{Chao Zhang}.} \bibinfo{year}{2023}\natexlab{b}.
\newblock \showarticletitle{Aegis: Mitigating Targeted Bit-flip Attacks against Deep Neural Networks}. In \bibinfo{booktitle}{\emph{32nd USENIX Security Symposium (USENIX Security 23)}}. \bibinfo{pages}{2329--2346}.
\newblock


\bibitem[Wang et~al\mbox{.}(2009)]%
        {wang2009pubchem}
\bibfield{author}{\bibinfo{person}{Yanli Wang}, \bibinfo{person}{Jewen Xiao}, \bibinfo{person}{Tugba~O Suzek}, \bibinfo{person}{Jian Zhang}, \bibinfo{person}{Jiyao Wang}, {and} \bibinfo{person}{Stephen~H Bryant}.} \bibinfo{year}{2009}\natexlab{}.
\newblock \showarticletitle{PubChem: a public information system for analyzing bioactivities of small molecules}.
\newblock \bibinfo{journal}{\emph{Nucleic acids research}}  \bibinfo{volume}{37} (\bibinfo{year}{2009}), \bibinfo{pages}{623--633}.
\newblock


\bibitem[Wang et~al\mbox{.}(2023a)]%
        {wang2023dynamic}
\bibfield{author}{\bibinfo{person}{Zhiqiong Wang}, \bibinfo{person}{Zican Lin}, \bibinfo{person}{Shuo Li}, \bibinfo{person}{Yibo Wang}, \bibinfo{person}{Weiying Zhong}, \bibinfo{person}{Xinlei Wang}, {and} \bibinfo{person}{Junchang Xin}.} \bibinfo{year}{2023}\natexlab{a}.
\newblock \showarticletitle{Dynamic Multi-Task Graph Isomorphism Network for Classification of Alzheimer’s Disease}.
\newblock \bibinfo{journal}{\emph{Applied Sciences}} \bibinfo{volume}{13}, \bibinfo{number}{14} (\bibinfo{year}{2023}), \bibinfo{pages}{8433}.
\newblock


\bibitem[Welling and Kipf(2016)]%
        {welling2016semi}
\bibfield{author}{\bibinfo{person}{Max Welling} {and} \bibinfo{person}{Thomas~N Kipf}.} \bibinfo{year}{2016}\natexlab{}.
\newblock \showarticletitle{Semi-supervised classification with graph convolutional networks}. In \bibinfo{booktitle}{\emph{4th International Conference on Learning Representations}}.
\newblock


\bibitem[Wu et~al\mbox{.}(2023)]%
        {wu2023securing}
\bibfield{author}{\bibinfo{person}{Bang Wu}, \bibinfo{person}{Xingliang Yuan}, \bibinfo{person}{Shuo Wang}, \bibinfo{person}{Qi Li}, \bibinfo{person}{Minhui Xue}, {and} \bibinfo{person}{Shirui Pan}.} \bibinfo{year}{2023}\natexlab{}.
\newblock \showarticletitle{Securing Graph Neural Networks in MLaaS: A Comprehensive Realization of Query-based Integrity Verification}.
\newblock \bibinfo{journal}{\emph{CoRR}}  \bibinfo{volume}{abs/2312.07870} (\bibinfo{year}{2023}).
\newblock


\bibitem[Wu et~al\mbox{.}(2022)]%
        {wu2022graph}
\bibfield{author}{\bibinfo{person}{Lingfei Wu}, \bibinfo{person}{Peng Cui}, \bibinfo{person}{Jian Pei}, \bibinfo{person}{Liang Zhao}, {and} \bibinfo{person}{Le Song}.} \bibinfo{year}{2022}\natexlab{}.
\newblock \bibinfo{booktitle}{\emph{Graph Neural Networks: Foundations, Frontiers, and Applications}}.
\newblock


\bibitem[Wu et~al\mbox{.}(2018)]%
        {wu2018moleculenet}
\bibfield{author}{\bibinfo{person}{Zhenqin Wu}, \bibinfo{person}{Bharath Ramsundar}, \bibinfo{person}{Evan~N. Feinberg}, \bibinfo{person}{Joseph Gomes}, \bibinfo{person}{Caleb Geniesse}, \bibinfo{person}{Aneesh~S. Pappu}, \bibinfo{person}{Karl Leswing}, {and} \bibinfo{person}{Vijay Pande}.} \bibinfo{year}{2018}\natexlab{}.
\newblock \showarticletitle{MoleculeNet: a benchmark for molecular machine learning}.
\newblock \bibinfo{journal}{\emph{Chemical science}} \bibinfo{volume}{9}, \bibinfo{number}{2} (\bibinfo{year}{2018}), \bibinfo{pages}{513--530}.
\newblock


\bibitem[Xiong et~al\mbox{.}(2021)]%
        {xiong2021novo}
\bibfield{author}{\bibinfo{person}{Jiacheng Xiong}, \bibinfo{person}{Zhaoping Xiong}, \bibinfo{person}{Kaixian Chen}, \bibinfo{person}{Hualiang Jiang}, {and} \bibinfo{person}{Mingyue Zheng}.} \bibinfo{year}{2021}\natexlab{}.
\newblock \showarticletitle{Graph neural networks for automated de novo drug design}.
\newblock \bibinfo{journal}{\emph{Drug Discovery Today}} \bibinfo{volume}{26}, \bibinfo{number}{6} (\bibinfo{year}{2021}), \bibinfo{pages}{1382--1393}.
\newblock


\bibitem[Xu et~al\mbox{.}(2020)]%
        {xu2020adversarial}
\bibfield{author}{\bibinfo{person}{Han Xu}, \bibinfo{person}{Yao Ma}, \bibinfo{person}{Hao-Chen Liu}, \bibinfo{person}{Debayan Deb}, \bibinfo{person}{Hui Liu}, \bibinfo{person}{Ji-Liang Tang}, {and} \bibinfo{person}{Anil~K Jain}.} \bibinfo{year}{2020}\natexlab{}.
\newblock \showarticletitle{Adversarial attacks and defenses in images, graphs and text: A review}.
\newblock \bibinfo{journal}{\emph{International Journal of Automation and Computing}}  \bibinfo{volume}{17} (\bibinfo{year}{2020}), \bibinfo{pages}{151--178}.
\newblock


\bibitem[Xu et~al\mbox{.}(2021)]%
        {xu2021survey}
\bibfield{author}{\bibinfo{person}{Jingjing Xu}, \bibinfo{person}{Wangchunshu Zhou}, \bibinfo{person}{Zhiyi Fu}, \bibinfo{person}{Hao Zhou}, {and} \bibinfo{person}{Lei Li}.} \bibinfo{year}{2021}\natexlab{}.
\newblock \showarticletitle{A survey on green deep learning}.
\newblock \bibinfo{journal}{\emph{CoRR}}  \bibinfo{volume}{abs/2111.05193} (\bibinfo{year}{2021}).
\newblock


\bibitem[Xu et~al\mbox{.}(2019)]%
        {leskovec2019powerful}
\bibfield{author}{\bibinfo{person}{Keyulu Xu}, \bibinfo{person}{Weihua Hu}, \bibinfo{person}{Jure Leskovec}, {and} \bibinfo{person}{Stefanie Jegelka}.} \bibinfo{year}{2019}\natexlab{}.
\newblock \showarticletitle{How Powerful are Graph Neural Networks?}. In \bibinfo{booktitle}{\emph{7th International Conference on Learning Representations}}.
\newblock


\bibitem[Yan et~al\mbox{.}(2020)]%
        {yan2020cache}
\bibfield{author}{\bibinfo{person}{Mengjia Yan}, \bibinfo{person}{Christopher~W Fletcher}, {and} \bibinfo{person}{Josep Torrellas}.} \bibinfo{year}{2020}\natexlab{}.
\newblock \showarticletitle{Cache telepathy: Leveraging shared resource attacks to learn $\{$DNN$\}$ architectures}. In \bibinfo{booktitle}{\emph{29th USENIX Security Symposium (USENIX Security 20)}}. \bibinfo{pages}{2003--2020}.
\newblock


\bibitem[Yang et~al\mbox{.}(2022)]%
        {yang2022identification}
\bibfield{author}{\bibinfo{person}{Sihong Yang}, \bibinfo{person}{Dezhi Jin}, \bibinfo{person}{Jun Liu}, {and} \bibinfo{person}{Ye He}.} \bibinfo{year}{2022}\natexlab{}.
\newblock \showarticletitle{Identification of Young High-Functioning Autism Individuals Based on Functional Connectome Using Graph Isomorphism Network: A Pilot Study}.
\newblock \bibinfo{journal}{\emph{Brain Sciences}} \bibinfo{volume}{12}, \bibinfo{number}{7} (\bibinfo{year}{2022}), \bibinfo{pages}{883}.
\newblock


\bibitem[Yao et~al\mbox{.}(2020)]%
        {yao2020deephammer}
\bibfield{author}{\bibinfo{person}{Fan Yao}, \bibinfo{person}{Adnan~Siraj Rakin}, {and} \bibinfo{person}{Deliang Fan}.} \bibinfo{year}{2020}\natexlab{}.
\newblock \showarticletitle{{DeepHammer}: Depleting the intelligence of deep neural networks through targeted chain of bit flips}. In \bibinfo{booktitle}{\emph{29th USENIX Security Symposium (USENIX Security 20)}}. \bibinfo{pages}{1463--1480}.
\newblock


\bibitem[Yao et~al\mbox{.}(2022)]%
        {yao2022edge}
\bibfield{author}{\bibinfo{person}{Jiangchao Yao}, \bibinfo{person}{Shengyu Zhang}, \bibinfo{person}{Yang Yao}, \bibinfo{person}{Feng Wang}, \bibinfo{person}{Jianxin Ma}, \bibinfo{person}{Jianwei Zhang}, \bibinfo{person}{Yunfei Chu}, \bibinfo{person}{Luo Ji}, \bibinfo{person}{Kunyang Jia}, \bibinfo{person}{Tao Shen}, {et~al\mbox{.}}} \bibinfo{year}{2022}\natexlab{}.
\newblock \showarticletitle{Edge-cloud polarization and collaboration: A comprehensive survey for ai}.
\newblock \bibinfo{journal}{\emph{IEEE Transactions on Knowledge and Data Engineering}} \bibinfo{volume}{35}, \bibinfo{number}{7} (\bibinfo{year}{2022}), \bibinfo{pages}{6866--6886}.
\newblock


\bibitem[Zaheer et~al\mbox{.}(2017)]%
        {ZaheerKRPSS17}
\bibfield{author}{\bibinfo{person}{Manzil Zaheer}, \bibinfo{person}{Satwik Kottur}, \bibinfo{person}{Siamak Ravanbakhsh}, \bibinfo{person}{Barnab{\'{a}}s P{\'{o}}czos}, \bibinfo{person}{Ruslan Salakhutdinov}, {and} \bibinfo{person}{Alexander~J. Smola}.} \bibinfo{year}{2017}\natexlab{}.
\newblock \showarticletitle{Deep Sets}. In \bibinfo{booktitle}{\emph{Advances in Neural Information Processing Systems 30}}. \bibinfo{pages}{3391--3401}.
\newblock


\bibitem[Zhang and Chung(2021)]%
        {zhang2021medq}
\bibfield{author}{\bibinfo{person}{Rongzhao Zhang} {and} \bibinfo{person}{Albert C.~S. Chung}.} \bibinfo{year}{2021}\natexlab{}.
\newblock \showarticletitle{MedQ: Lossless ultra-low-bit neural network quantization for medical image segmentation}.
\newblock \bibinfo{journal}{\emph{Medical Image Analysis}}  \bibinfo{volume}{73} (\bibinfo{year}{2021}), \bibinfo{pages}{102200}.
\newblock


\bibitem[Zhang et~al\mbox{.}(2022)]%
        {zhang2022unsupervised}
\bibfield{author}{\bibinfo{person}{Sixiao Zhang}, \bibinfo{person}{Hongxu Chen}, \bibinfo{person}{Xiangguo Sun}, \bibinfo{person}{Yicong Li}, {and} \bibinfo{person}{Guandong Xu}.} \bibinfo{year}{2022}\natexlab{}.
\newblock \showarticletitle{Unsupervised graph poisoning attack via contrastive loss back-propagation}. In \bibinfo{booktitle}{\emph{Proceedings of the ACM Web Conference 2022}}. \bibinfo{pages}{1322--1330}.
\newblock


\bibitem[Zhu et~al\mbox{.}(2023)]%
        {zhu2023rm}
\bibfield{author}{\bibinfo{person}{Zeyu Zhu}, \bibinfo{person}{Fanrong Li}, \bibinfo{person}{Zitao Mo}, \bibinfo{person}{Qinghao Hu}, \bibinfo{person}{Gang Li}, \bibinfo{person}{Zejian Liu}, \bibinfo{person}{Xiaoyao Liang}, {and} \bibinfo{person}{Jian Cheng}.} \bibinfo{year}{2023}\natexlab{}.
\newblock \showarticletitle{$A^2$Q: Aggregation-Aware Quantization for Graph Neural Networks}. In \bibinfo{booktitle}{\emph{11th International Conference on Learning Representations}}.
\newblock


\bibitem[Zopf(2022)]%
        {Zopf22a}
\bibfield{author}{\bibinfo{person}{Markus Zopf}.} \bibinfo{year}{2022}\natexlab{}.
\newblock \showarticletitle{1-WL Expressiveness Is (Almost) All You Need}. In \bibinfo{booktitle}{\emph{International Joint Conference on Neural Networks, {IJCNN}}}. \bibinfo{pages}{1--8}.
\newblock


\bibitem[Z\"{u}gner et~al\mbox{.}(2018)]%
        {zunger2018adversarial}
\bibfield{author}{\bibinfo{person}{Daniel Z\"{u}gner}, \bibinfo{person}{Amir Akbarnejad}, {and} \bibinfo{person}{Stephan G\"{u}nnemann}.} \bibinfo{year}{2018}\natexlab{}.
\newblock \showarticletitle{Adversarial Attacks on Neural Networks for Graph Data}. In \bibinfo{booktitle}{\emph{Proceedings of the 24th ACM SIGKDD International Conference on Knowledge Discovery \& Data Mining}}. \bibinfo{pages}{2847–2856}.
\newblock


\bibitem[Zügner and Günnemann(2019)]%
        {zugner2019adversarial}
\bibfield{author}{\bibinfo{person}{Daniel Zügner} {and} \bibinfo{person}{Stephan Günnemann}.} \bibinfo{year}{2019}\natexlab{}.
\newblock \showarticletitle{Adversarial Attacks on Graph Neural Networks via Meta Learning}. In \bibinfo{booktitle}{\emph{7th International Conference on Learning Representations}}.
\newblock


\end{thebibliography}

\newpage
\appendix

\appendix
\section{Proof of Proposition 3.3}
\begin{proof} %
 We first prove by induction that the statement Equation~\eqref{eq:statementA} implies that there is a 1-to-1 correspondence between $\mathcal{J}_{i}$ and the isomorphism types of unfolding trees of height $i$, denoted by $\mathcal{T}_i$, for all $i \in \{0,\dots, k\}$. In the base case $i=0$, there is a single unfolding tree in $\mathcal{T}_0$ consisting of a single node. The uniform initialization $\mathcal{J}_0$ satisfies the requirement. Assume that $\varphi$ is a bijection between $\mathcal{J}_{i}$ and $\mathcal{T}_i$, then the statement~\eqref{eq:statementA} together with the permutation-invariance guarantees that $f^{(i+1)}(A,\mathcal{A})=f^{(i+1)}(B,\mathcal{B})$ if and only if $A=B$ and $\mathcal{A}=\mathcal{B}$. Hence, $\multiset{\varphi(a) \mid a \in \mathcal{A}} = \multiset{\varphi(b) \mid b \in \mathcal{B}}$, which uniquely determines an unfolding tree in $\mathcal{T}_{i+1}$ according to Definition~\ref{def:unfolding}.
 Vice versa, unfolding trees with different subtrees lead to distinguishable multisets.
 The result follows by Lemma~\ref{lem:wlc} and the 1-to-1 correspondence shown above at layer $k$.
\end{proof}

\section{Proof of Proposition 3.5}
\begin{proof}
\LK{
Assume
$F^{(k)}_G(G_i)=F^{(k)}_G(G_j)$ but $C_l^{(k)}(G_i) \neq C_l^{(k)}(G_j)$.
Then
$f_G(\multiset{F^{(k)}(v) \mid v \in V(G_i)}) = f_G(\multiset{F^{(k)}(v) \mid v \in V(G_j)})$
but
$\multiset{c_l^{(k)}(v) \mid v \in V(G_i)} \neq \multiset{c_l^{(k)}(v) \mid v \in V(G_j)}$. 
Because per construction $c_l^{(k)}(u)=c_l^{(k)}(v) \Longleftrightarrow F^{(k)}(u) = F^{(k)}(v)$ it follows that $ \mathcal{A} \not{=} \mathcal{B}$, contradicting} Equation~\eqref{eq:statementA2}.
\end{proof}

\section{A motivating case study}
Our preliminary case study summarized in tab.~\ref{bnnfia_prelim} indicates a significant vulnerability of GNNs used in community-based tasks on graphs with strong homophily~\citep{rossi2020proximity} to malicious BFAs such as PBFA, since it suggests a quantized GNN can be degraded so severely by an extremely small number of bit flips---relative to the network's attack surface---that it produces basically random output. In our case study, a GNN's output on the community-based tasks is random if its test accuracy drops below 14.3\% (= 1/7) on \texttt{Cora}'s 7-class node classification task or below 16.7\% (= 1/6) on \texttt{CiteSeer}'s 6-class node classification task. Our case study shows that this is consistently the case due to the PBFA adapted from~\citep{rakin2019bitflip} for all %
community-based architecture-dataset combinations examined. The number of bit flips required for completely degrading a GNN in a community-based task is remarkably small: tab.~\ref{bnnfia_prelim} shows that on average, the adapted PBFA flipped only %
0.0004\%of the total number of bits of the quantized GNNs' parameters. Regarding random bit flips (\textbf{RBFA}), the results of our case study are consistent with results obtained for full-precision GNNs~\citep{jiao2022assessing} in that they demonstrate a relatively strong resilience of GNNs against such random perturbations. 

On structural tasks requiring high structural expressivity on graphs with weak/low homophily as is typical in molecular, chemical, and protein networks~\citep{rossi2020proximity} which are common in, e.g., drug development, PBFA is much less effective and degrades the network comparable to random bit flips. On the tasks requiring high structural expressivity in tab.~\ref{bnnfia_prelim}, a GNN’s output is random if its test AUROC drops to 0.5. We found that on the \texttt{ogbg-moltoxcast} dataset, PBFA could not significantly degrade the network even after 2662 flips and that on \texttt{ogbg-molhiv}, 0,0063\% of the total number of bits of the quantized GNNs' parameters had to be flipped by PBFA before the GNN's output was degraded to random output, which constitutes a 15.75 times increase compared to the community-based tasks. This increased resilience of GNNs trained on tasks requiring high structural expressivity compared to community-based tasks cannot be explained entirely by the higher number of GNN parameters found in the evaluated tasks requiring high structural expressivity, which mostly stems from the MLPs employed in GIN: The increase in required flips for PBFA to entirely degrade the network on the task requiring high structural expressivity is up to 2 orders of magnitudes larger compared to the community-based task, while the increase in the attack surface is at most 1 order of magnitude larger. 
Based on these observations, our work focuses on such tasks requiring high structural expressivity, which are typically solved by GIN.
\begin{table*}[]
\centering
\caption{\label{bnnfia_prelim}%
Preliminary case study illustrating the general vulnerability of GNNs to PBFA -- pre- and post-attack mean of 10 runs of top-1 test accuracy (community) or AUROC (structure) of \texttt{INT8} quantized representative GNN architecture (GCN~\citep{wu2022graph} with 3 layers, GAT~\citep{wu2022graph} with 2 layers, GIN~\citep{leskovec2019powerful} with 5 layers) and dataset combinations (GCN on \texttt{Cora}, GAT on \texttt{CiteSeer}, GIN on \texttt{ogbg-mol}) baseline without BFAs; after PBFA~\citep{rakin2019bitflip} adapted to GNNs; 
after random bit flips (RBFA); total bit count of all model parameters (attack surface) in millions. 
}
\begin{tabular}{@{}lllllllllll@{}}
\toprule
 & \multicolumn{5}{c}{\bf COMMUNITY} & \multicolumn{5}{c}{ \bf STRUCTURAL} \\ %
 \hline \\
\bf Attack & \bf Dataset & \bf Pre & \bf Post & \bf Flips & \bf Bits & \bf Dataset & \bf Pre & \bf Post & \bf Flips & \bf Bits \\ %
RBFA & \texttt{Cora}-GCN & 0.77 & 0.74 & 63 & 1.6M & \texttt{ogbg-molhiv}-GIN & 0.71 & 0.53 & 953 & 15.1M \\
PBFA & \texttt{Cora}-GCN & 0.77 & 0.12 & 9 & 1.6M & \texttt{ogbg-molhiv}-GIN & 0.71 & 0.50 & 953 & 15.1M \\
RBFA & \texttt{CiteSeer}-GAT & 0.58 & 0.48 & 63 & 3.8M & \texttt{ogbg-moltoxcast}-GIN & 0.58 & 0.58 & 2662 & 16.6M \\
PBFA & \texttt{CiteSeer}-GAT & 0.58 & 0.14 & 10 & 3.8M & \texttt{ogbg-moltoxcast}-GIN & 0.58 & 0.57 & 2662 & 16.6M \\ 
\bottomrule
\end{tabular}
\end{table*}

\section{Progressive bit flip attack}
The PBFA on CNN weights is an attack methodology that can crush a CNN by maliciously flipping minimal numbers of bits within its weight storage memory (i.e., DRAM). It was first introduced as an untargeted attack~\citep{rakin2019bitflip}. PBFA operates on integer quantized CNNs (as described above) and seeks to optimize Equation~\eqref{eq:bfa_optimization2}.
\begin{equation}
    \begin{aligned}
            \max_{\{ \widehat{\mathbf{W}}_{q}^{(l)} \}} \quad & \mathcal{L}\Big (\Phi \big( \mathbf{X}; \; \{\widehat{\textbf{W}}_{q}^{(l)}\}_{l=1}^{L} \big), \; \mathbf{t} \Big) - \mathcal{L}\Big (\Phi \big( \mathbf{X}; \; \{\textbf{W}_{q}^{(l)}\}_{l=1}^{L} \big), \; \mathbf{t} \Big) \\
            \textrm{ s.t.} 
            \quad & \sum_{l=1}^L \mathcal{D}(\widehat{\mathbf{W}}_{q}^{(l)}, \; \mathbf{W}_{q}^{(l)}) \; \in \; \{ 0,1, \dots, N_b \} 
            \\
    \end{aligned}
    \label{eq:bfa_optimization2}
\end{equation}
where $\mathbf{X}$ and $\mathbf{t}$ are input batch and target vector, $\mathcal{L}$ is a loss function, $f$ is a neural network, $L$ is the number of layers and $\widehat{\mathbf{W}}_{q}^{(l)}, \mathbf{W}_{q}^{(l)}$ are the perturbed and unperturbed integer quantized weights (stored in two's complement) of layer $l$. In the original work by~\citep{rakin2019bitflip}, the function $\mathcal{L}$ used is the same loss originally used during network training. $\mathcal{D}(\widehat{\mathbf{W}}_{q}^{(l)}, \mathbf{W}_{q}^{(l)})$ represents the Hamming distance between clean- and perturbed-binary weight tensor, and $N_b$ represents the maximum Hamming distance allowed through the entire
CNN. 

The attack is executed by flipping the bits along its gradient ascending direction w.r.t.~the loss of CNN. That is, using the $N_q$-bits binary representation $\mathbf{b} = [b_{N_q-1}, \dots, b_{0}]$ of weights $w \in \mathbf{W}_{q,l}$, first the gradients of $\mathbf{b}$ w.r.t. to inference loss $\mathcal{L}$ are computed
\begin{equation}
    \nabla_{\mathbf{b}} \mathcal{L} \left [ \frac{\partial\mathcal{L}}{\partial b_{N_q-1}}, \dots, \frac{\partial\mathcal{L}}{\partial b_{0}} \right ]
\end{equation}
and then the perturbed bits are computed via $\mathbf{m} = \mathbf{b} \oplus (\textup{sign}(\nabla_{\mathbf{b}}\mathcal{L})/2+0.5) $ and $\widehat{\mathbf{b}} = \mathbf{b} \oplus \mathbf{m}$, where $\oplus$ denotes the bitwise \texttt{xor} operator.

To improve efficiency over iterating through each bit of the entire CNN, the authors employ a method called progressive bit search (\textbf{PBS}). As noted earlier, we refer to this BFA variant employing PBS as Progressive BFA or PBFA. In PBS, at each iteration of the attack (to which we synonymously refer as \emph{attack run}), in a first step for each layer {$l \in \left [ 0,  L \right ]$, %
the $n_b$ most vulnerable bits in $\widehat{\textbf{W}}_{q}^{(l)}$ are identified through gradient ranking (in-layer search). That is, regarding input batch $\mathbf{X}$ and target vector $\mathbf{t}$, inference and backpropagation are performed successively to calculate the gradients of bits w.r.t.~the inference loss and the bits are ranked by the absolute values of their gradients $\partial \mathcal{L}/ \partial b$. In a second step, after the most vulnerable bit per layer is identified, the gradients are ranked across all layers s.t.~the most vulnerable bit in the entire CNN is found (cross-layer search) and flipped.

Should an iteration of PBS not yield an attack solution, which can be the case if no single bit flip improves the optimization goal given in Equation~\eqref{eq:bfa_optimization2}, PBS is executed again and evaluates increasing combinations of $2$ or more bit flips.

\section{Dataset descriptions}
The task associated with the \texttt{ogb-molhiv} dataset is to predict whether a certain molecule structure inhibits human immunodeficiency virus (HIV) or not. In the larger \texttt{ogb-molpcba} dataset each graph represents a molecule, where nodes are atoms, and edges are chemical bonds, and the task is to predict 128 different biological activities (inactive/active). The \texttt{ogb-moltox21} dataset contains data with qualitative toxicity measurements on 12 biological targets. The \texttt{ogbg-toxcast} dataset is another toxicity related dataset. The \texttt{obgbg-molbace} dataset is a biochemical single task binary classification (inhibition of human $\beta$-secretase 1 (BACE-1)) dataset. The \texttt{ogbg-molmuv} dataset is a subset of PubChem BioAssay commonly used for evaluation of virtual screening techniques. The \texttt{COLLAB} dataset consist of ego-networks extracted from scientific collaboration networks. In these datasets, each ego-network represents a researcher, and the objective is to forecast their specific area of research, such as high energy physics, condensed matter physics, or astrophysics. \texttt{GITHUB_STARGAZERS} contains graphs depicting GitHub users' social networks, divided based on their interactions with popular machine learning and web development repositories.
~\citep{hu2020open, gao2022patchgt, suresh2021adversarial, wang2009pubchem, morris2020tudataset}. In tab.~\ref{tab:ogb_dataset}, an overview of the datasets’ structure is provided.

\begin{table*}%
\centering
\caption{\label{tab:ogb_dataset} Overview of the eight real-world benchmark datasets from OGB \citep{hu2020open} and TUDataset~\citep{morris2020tudataset} that are used with number of graphs, average of number of nodes and edges and recommended performance metric.%
    }
\begin{tabular}{@{}lllllll@{}}
\toprule
\bf DATASET & \bf GRAPHS & \bf NODES & \bf EDGES & \bf TASKS & \bf METRIC & \bf TYPE\\ %
\hline \\
\texttt{ogbg-molpcba} & 437929 & 26.0 & 28.1 & 128 & AP & Binary Multi-Task \\
\texttt{ogbg-molmuv} & 93087 & 24.2 & 26.3 & 17 & AP & Binary Multi-Task \\
\texttt{ogbg-molhiv} & 41127 & 25.5 & 27.5 & 1 & AUROC & Binary Single-Task \\
\texttt{ogbg-moltoxcast} & 8576 & 18.5 & 19.3 & 617 & AUROC & Binary Multi-Task \\
\texttt{ogbg-moltox21} & 7831 & 18.6 & 19.3 & 12 & AUROC & Binary Multi-Task \\
\texttt{ogbg-molbace} & 1513 & 34.1 & 36.9 & 1 & AUROC & Binary Single-Task \\ %
\texttt{GITHUB_STARGAZERS} & 12725 & 391.4 & 456.9 & 1 & ACC & Binary Single-Task \\ %
\texttt{COLLAB} & 5000 & 74.5 & 4914.4 & 3 & ACC & Multi Class \\ \bottomrule
\end{tabular}
\end{table*}

\section{Ablation Study}
\subsection{Loss functions}
Figure~\ref{fig:res5} illustrates the results if L1 loss is used instead of KL loss in the multi-task binary classification setting. As can be seen from Figure~\ref{fig:res5}, IBFA1/2 both fail to outperform PBFA on the multi-task binary classification datasets if L1 loss is used instead of KL loss, which is in line with our analytical results in Section~\ref{par:loss}. As in Figure~\ref{fig:res4}, 1\% subset sampling was used for IBFA to accelerate the experiments. Experiments on \texttt{obg-molbace}, \texttt{ogbg-molhiv} and \texttt{GITHUB_STARGAZERS} are not included in this experiment as we already used L1 loss in our original experiments with these datasets.
\begin{figure*}%
    \centering %
    \includegraphics[width=2.0\columnwidth, trim={0.25cm 0.075cm 0.25cm 0.250cm}, clip]{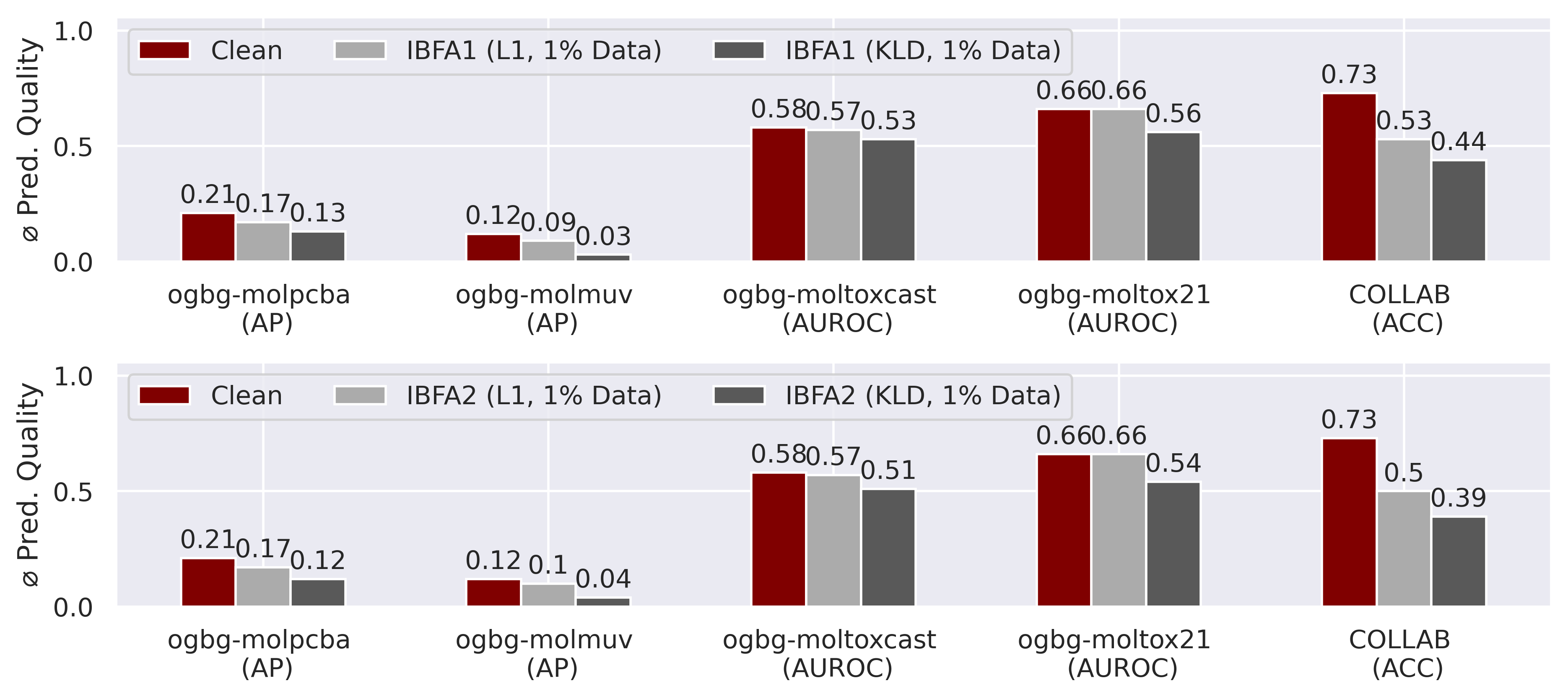} 
    \caption{\label{fig:res5} %
    Pre- \LK{(clean)} and post-attack test quality metrics AP, AUROC or ACC for IBFA with different loss functions (L1 vs. KL loss, IBFA selection from 1\% random subset) on a %
    5-layer GIN 
    trained on 4 \texttt{ogbg-mol} %
    multi-task binary classification datasets and TUDataset \texttt{COLLAB}, number of bit flips, averages of 10 runs. IBFA1 in top row, IBFA2 in bottom row. %
    }
\end{figure*} 
\subsection{Layer preferences}
In order to investigate the attack strategies employed by the evaluated attacks PBFA, IBFA1, and IBFA2, we recorded the probabilities associated with an attack's selection of a specific component within the evaluated 5-layer GIN (refer to Figure~\ref{fig:res3}). RBFA was configured to exhibit a random and uniform distribution of bit flips across the layers and is not shown in Figure~\ref{fig:res3}. On the other hand, PBFA and IBFA1/2 exhibit distinct and characteristic patterns in terms of layer selection. PBFA typically confines bit flips to only 2 out of the 5 layers and, in line with~\citet{hector2022evaluating}'s findings for CNNs, displays a preference for layers near the input layer, while IBFA1/2 targets bit flips in at least 4 layers across the entire model, with the majority of flips occurring in the learnable aggregation functions of the network, namely MLP1-4. The variations in layer selection observed in IBFA1/2 support our hypotheses: a) introducing non-injectivity into a single layer alone is insufficient, necessitating an attack on the overall expressivity of GIN, and b) IBFA1/2 effectively targets the learnable neighborhood aggregation functions.
\begin{figure*}%
    \centering %
    \includegraphics[width=1.5\columnwidth, trim={0.25cm 0.075cm 0.25cm 0.250cm}, clip]{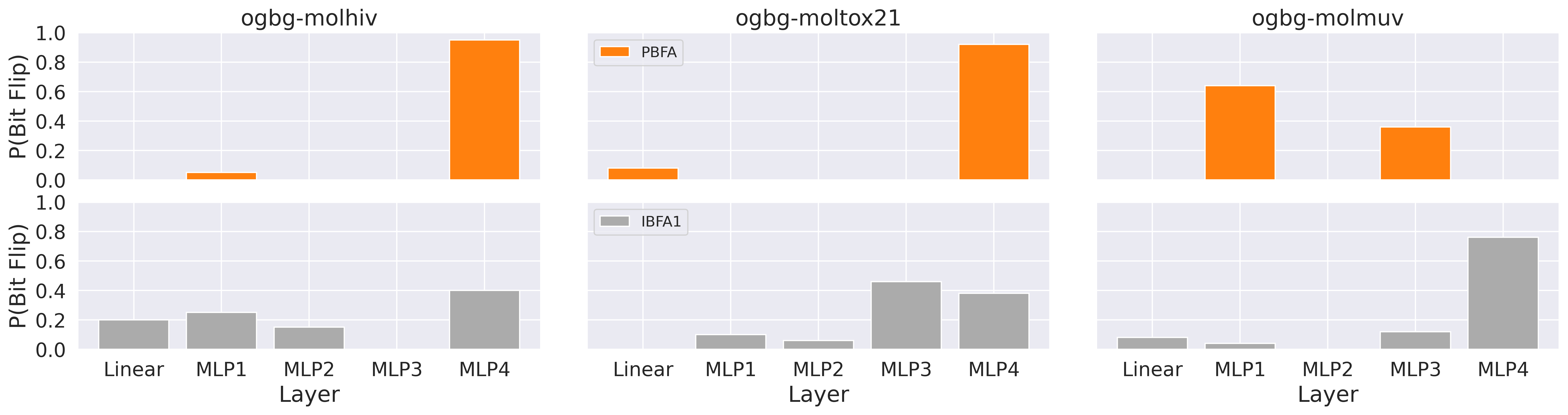} 
    \caption{\label{fig:res3} %
    Probability of a component of a 5-layer GIN %
    trained on 3 \texttt{ogbg-mol} %
    datasets %
    being selected for a bit flip by one of the 3 evaluated attacks, averaged over 10 runs. MLP1-4 denote the learnable neighborhood aggregation functions used in GIN, Linear denotes the linear output layer used for graph classification.}
\end{figure*} 

\section{Other GNN architectures}%
Motivated by our case study's findings as well as to obtain a first impression on IBFAs capability to generalize beyond GIN to other architectures, we repeated several experiments described in \NKT{Section}~\ref{sec:experiments} using Graph Convolutional Network (GCN)~\NKT{\citep{welling2016semi}} instead of GIN. In GCN, an element-wise mean pooling approach is employed for the $\Comb$ operation, and the steps of $\Agg$ and $\Comb$ are integrated in the following manner~\NKT{\citep{leskovec2019powerful}}:
\begin{equation}
    \mathbf{h}_v^{(k)} = \ReLU \left ( \mathbf{W}^{(k)} \cdot \Mean \left \{ \mathbf{h}_u^{(k-1)} \mid u \in N(v) \cup \{v\} \right \} \right )
    \label{eq:defgcn}
\end{equation}
The mean aggregator used by GCN is not an injective multiset function and thus GCN's expressive power is limited~\NKT{\citep{leskovec2019powerful}}.
\begin{figure*}%
    \centering %
    \includegraphics[width=1.75\columnwidth, trim={0.25cm 0.075cm 0.25cm 0.250cm}, clip]{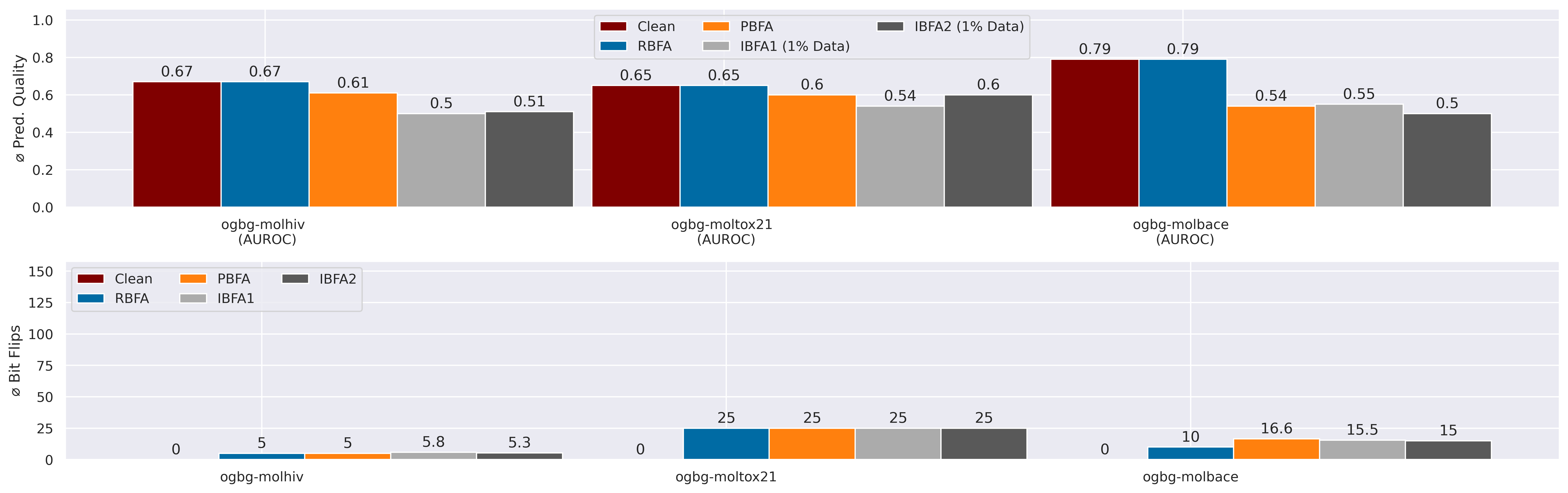} 
    \caption{\label{fig:res6} %
    Pre-(clean) and post-attack test quality metrics AP or AUROC for RBFA, PBFA, IBFA1/2 (selection from 1\% random subset),  on a %
    5-layer GCN 
    trained on 3 \texttt{ogbg-mol} datasets, averages of 10 runs.%
    }
\end{figure*} 
Figure~\ref{fig:res6} illustrates results for a 5-layer GCN trained on 4 \texttt{ogbg-mol} datasets. As can be seen from Figure~\ref{fig:res6}, IBFA1/2 outperforms or is on par with PBFA for GCN. As in Figure~\ref{fig:res4}, 1\% subset sampling was used for IBFA to expedite the experiments. Although not exhaustive, experiments in Figure~\ref{fig:res6} provide empirical support for our method’s ability to extend beyond its initial target architecture, GIN.

\end{document}